\documentclass{article}

\usepackage{graphicx}
\usepackage{amssymb,amsthm,amsmath,bm}
\usepackage{xcolor,paralist,hyperref,url,titlesec,fancyhdr,etoolbox}
\usepackage{caption,subcaption,multirow,tikz,booktabs}
\usepackage{authblk}
\usepackage{natbib}
\setcitestyle{authoryear}
\usepackage[a4paper, margin=2.9cm]{geometry}

\definecolor{bluecite}{HTML}{0875b7}
\hypersetup{
  colorlinks=true,
  citecolor=bluecite,
  linkcolor=magenta
}

\begin{document}

\title{\textbf{Multi-Agent Reinforcement Learning with Selective State-Space Models}}

\author{Jemma Daniel\thanks{Corresponding author: \href{mailto:j.daniel@instadeep.com}{\texttt{j.daniel@instadeep.com}}}}
\author{Ruan de Kock}
\author{Louay Ben Nessir}
\author{Sasha Abramowitz}
\author{Omayma Mahjoub}
\author{Wiem Khlifi}
\author{Claude Formanek}
\author{Arnu Pretorius}
\affil{InstaDeep}
\date{}

\maketitle


\begin{abstract}
The Transformer model has demonstrated success across a wide range of domains, including in Multi-Agent Reinforcement Learning (MARL) where the Multi-Agent Transformer (MAT) has emerged as a leading algorithm in the field.
However, a significant drawback of Transformer models is their quadratic computational complexity relative to input size, making them computationally expensive when scaling to larger inputs.
This limitation restricts MAT’s scalability in environments with many agents.
Recently, State-Space Models (SSMs) have gained attention due to their computational efficiency, but their application in MARL remains unexplored.
In this work, we investigate the use of Mamba, a recent SSM, in MARL and assess whether it can match the performance of MAT while providing significant improvements in efficiency.
We introduce a modified version of MAT that incorporates standard and bi-directional Mamba blocks, as well as a novel ‘cross-attention’ Mamba block.
Extensive testing shows that our Multi-Agent Mamba (MAM) matches the performance of MAT across multiple standard multi-agent environments, while offering superior scalability to larger agent scenarios.
This is significant for the MARL community, because it indicates that SSMs could replace Transformers without compromising performance, whilst also supporting more effective scaling to higher numbers of agents.
Our project page is available at \url{https://sites.google.com/view/multi-agent-mamba}.
\end{abstract}


\section{Introduction}


\begin{figure}[t]
    \centering
    \begin{minipage}[b]{0.48\linewidth}
        \centering
        \includegraphics[width=\linewidth]{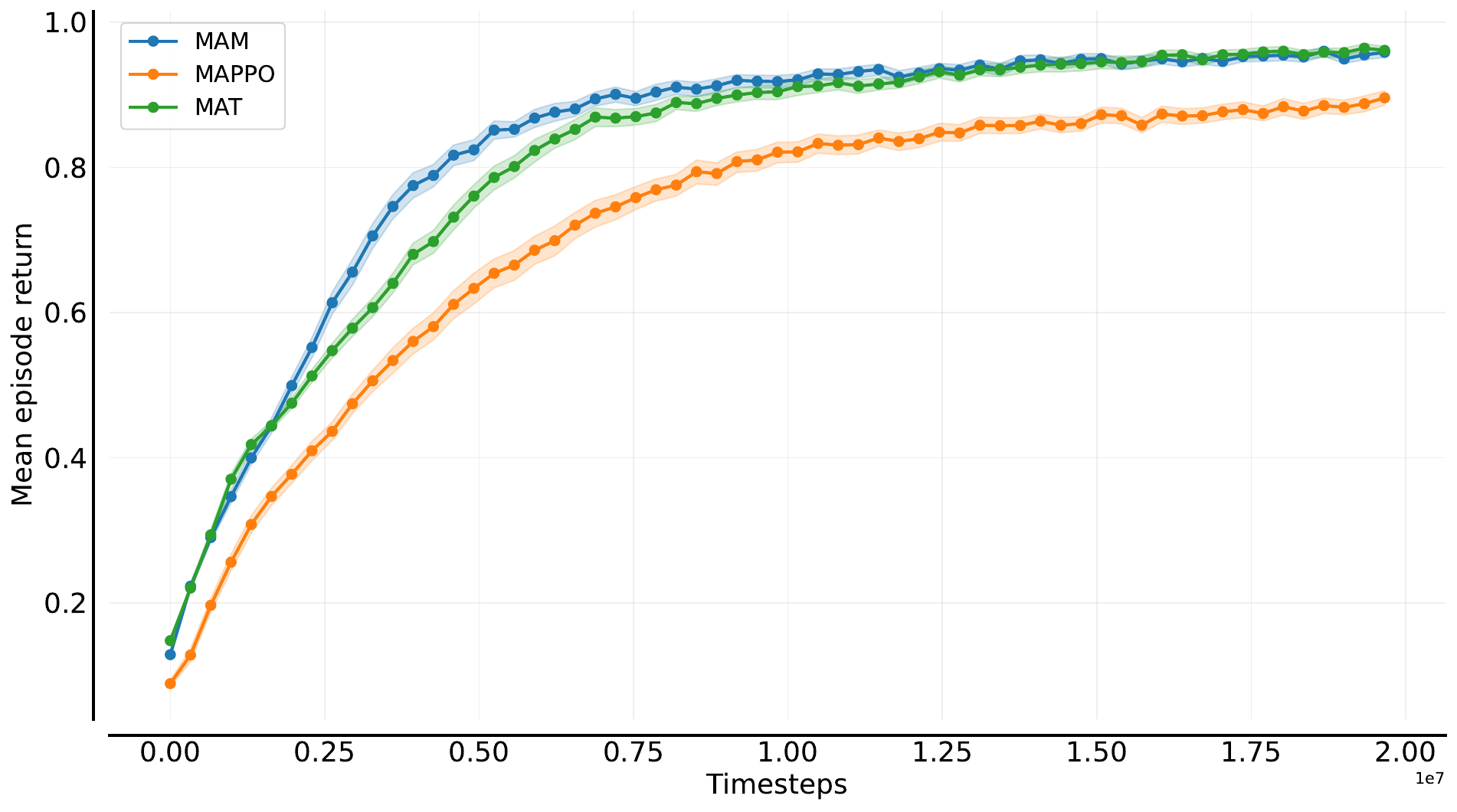}
        \caption{Normalised mean episode returns aggregated over all tasks and environments with 95\% confidence intervals for MAM, MAT, and MAPPO.
Results are obtained using ten independent seeds.
MAM matches the final performance of MAT, currently state-of-the-art in MARL, and exhibits greater sample efficiency.}
        \label{fig:all_envs_returns}
    \end{minipage}
    \hfill
    \begin{minipage}[b]{0.48\linewidth}
        \centering
        \includegraphics[width=\linewidth]{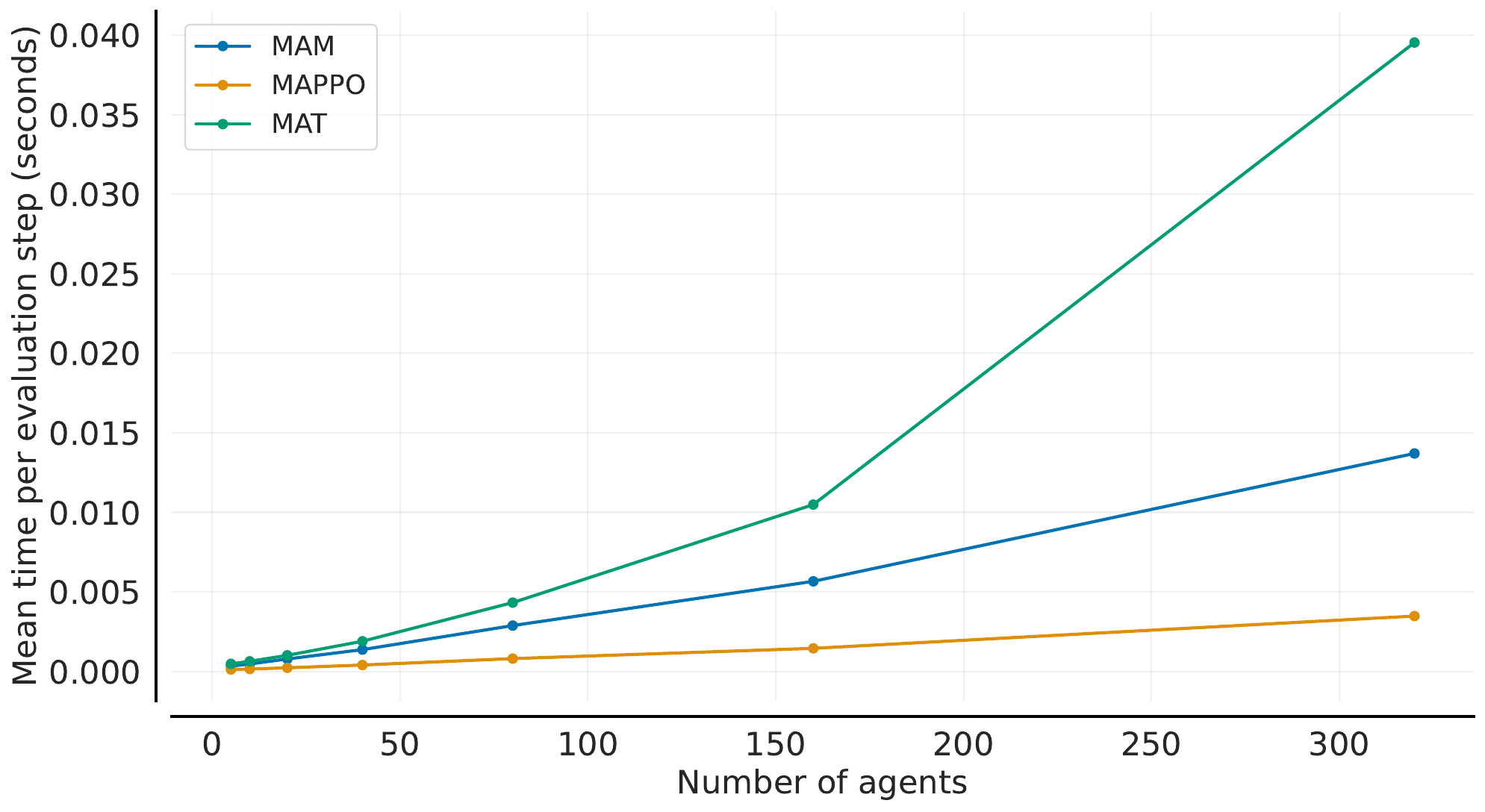}
        \vspace{0pt}
        \caption{Mean time (seconds) per evaluation step in \texttt{smacv2} tasks with increasing numbers of agents for MAM, MAT, and MAPPO.
The mean time per evaluation step for MAT increases approximately quadratically, while MAM and MAPPO scale linearly in the number of agents.}
        \label{fig:inference_sps_results}
    \end{minipage}
    \caption{Comparison of performance metrics for MAM, MAT, and MAPPO across various tasks.}
    \label{fig:comparison}
\end{figure}


Multi-Agent Reinforcement Learning (MARL) still faces significant challenges that must be overcome to unlock its full potential; one such challenge is the ability to scale to large numbers of agents while maintaining good performance.

The Multi-Agent Transformer (MAT) \citep{wen2022multiagentreinforcementlearningsequence} boasts state-of-the-art performance in online MARL.
MAT is an encoder-decoder framework that utilises the multi-agent advantage decomposition theorem \citep{kuba2022trustregionpolicyoptimisation} to reframe the challenge of joint policy optimisation.
It converts the problem into a sequential decision-making process, simplifying the search for optimal policies across multiple agents.
However, Transformers scale quadratically in sequence length \citep{vaswani2023attentionneed}.
This creates a computational bottleneck for MAT as the number of agents becomes large.

Recently, State-Space Models (SSMs) \citep{gu2022efficientlymodelinglongsequences,gupta2022diagonalstatespaceseffective,gu2021combiningrecurrentconvolutionalcontinuoustime,gupta2022diagonalstatespaceseffective,smith2023simplifiedstatespacelayers,fu2023hungryhungryhipposlanguage} have offered a solution to this drawback in the Transformer architecture, with the ability to scale linearly in the sequence length.
Of interest in this work is Mamba \citep{gu2024mambalineartimesequencemodeling}--a selective SSM which boasts fast inference and linear scaling in the sequence length whilst matching the performance of attention architectures in the natural language processing domain.
The innovations of the Mamba model are its input-dependent SSM parameters and its hardware-aware parallel algorithm in recurrent mode.

In this paper, we explore replacing attention in the MAT architecture with Mamba blocks.
We make use of both existing vanilla and bi-directional Mamba blocks, as well as a novel `cross-attention' Mamba block we designed to replace MAT's cross-attention.
We evaluate our novel architecture, which we call the Multi-Agent Mamba (MAM), on a wide range of well-known MARL benchmark environments and compare its performance to MAT.

Our core contributions can be summarised as follows:
\begin{itemize}
    \item We create an extension of the vanilla Mamba block which can be used as a cross-attention replacement.
    \item We replace the three different attention blocks in the MAT architecture with vanilla, bi-directional and cross-attentional Mamba blocks respectively.
    \item We empirically validate that MAM performs comparably to MAT on a wide range of MARL environments while offering significantly greater efficiency.
\end{itemize}



\section{Background}

In the following section we outline the necessary background for our work, including the MARL problem formulation, the multi-agent decomposition theorem, MAT and SSMs.


\subsection{\texorpdfstring{Problem Formulation: Cooperative Multi-Agent Reinforcement \\Learning}{Problem Formulation: Cooperative Multi-Agent Reinforcement Learning}}

Like the Markov decision process formalisation of single-agent reinforcement learning, MARL can be framed as the Markov game $\langle\mathcal{N},\bm{\mathcal{O}},\bm{\mathcal{A}},R,P,\gamma\rangle$ \citep{littman1994markovgamesasa}.
Denoting $\mathcal{N}=\{1,...,n\}$ as the set of agents, $\bm{\mathcal{O}}=\prod_{i=1}^{n}\mathcal{O}^i$ is the product of the agents' local observation spaces, and $\bm{\mathcal{A}}=\prod_{i=1}^n\mathcal{A}^i$ is the joint action space: a product of the agents' action spaces.
$R: \bm{\mathcal{O}}\times\bm{\mathcal{A}}\rightarrow[-R_\text{max},R_\text{max}]$ is the joint reward junction.
$P: \bm{\mathcal{O}}\times\bm{\mathcal{A}}\times\bm{\mathcal{O}}\rightarrow\mathbb{R}$ denotes the transition probability function, and $\gamma\in[0,1)$ is the discount factor.

In cooperative MARL a team of $n$ agents seeks to maximise their joint reward over time.
Based on their individual local observations $o_t^i\in\mathcal{O}^i$, each agent $i=1,...,n$ uses the $i^\text{th}$ component of the team's joint policy $\bm{\pi}$ to take an action $a_t^i$ at timestep $t\in\mathbb{N}$.
All agents take actions simultaneously, denoted jointly as $\mathbf{a}_t=(a^1_t,...,a^n_t)$, and the resulting reward $R(\mathbf{o}_t,\mathbf{a}_t)$ is allocated to all agents, where $\mathbf{o}_t=(o^1_t,...,o^n_t)$ is the joint observation.

\subsection{Multi-Agent Advantage Decomposition Theorem}

In the multi-agent setting, Multi-Agent PPO \citep{yu2022surprisingeffectivenessppocooperative} (MAPPO) has emerged as one of the leading algorithms across several benchmarks.
MAPPO extends PPO to MARL by conditioning the advantage function on the joint observations of all agents.
This situates MAPPO within the class of algorithms known as Centralised Training with Decentralised Execution (CTDE).
The policy gradient $\nabla_{\theta} J(\theta)$ in MAPPO is given by the following: 

\begin{align}
&\sum_{i=1}^{n} \sum_{t=0}^{T} \min \left( r_t^i(\theta) A_t^i(\mathbf{o_t}, \mathbf{a_t}), \, \text{clip}(r_t^i(\theta), 1 - \epsilon, 1 + \epsilon) A_t^i(\mathbf{o_t}, \mathbf{a_t}) \right),
\end{align}
where the clip operator restricts the input value to the interval $[1 - \epsilon, 1 + \epsilon]$, and
\begin{align}
\quad r_t^i(\theta) = \frac{\pi_{\theta_i}(a_t^i | o_t^i)}{\pi_{\theta_i}^{\text{old}}(a_t^i | o_t^i)}.
\end{align}

However, more recently it has been shown that there is a more principled way to implement a multi-agent version of PPO that has monotonic improvement guarantees.
HAPPO \citep{kuba2022trustregionpolicyoptimisation} leverages a novel result known as the multi-agent advantage decomposition theorem to further improve performance across a range of tasks.

The multi-agent advantage decomposition theorem breaks the joint advantage function up into a sum of the $n$ individual advantage functions, allowing for sequential agent policy updates \citep{kuba2022trustregionpolicyoptimisation}.
Formally, given a random permutation of agents $i_{1:n}$ and for any joint observation and joint action, the following equation always holds:
\begin{equation}
    A_\mathbf{\pi}^{i_{i:n}} = \sum_{m=1}^n A_\mathbf{\pi}^{i_m} \left( \mathbf{o},\mathbf{a}^{i_{1:m-1}},a^m \right).
\end{equation}
This theorem provides an intuitive approach: each agent sequentially selects its own action conditioned on the joint observation of all agents and the actions taken by preceding agents.
This sequential update scheme enjoys guaranteed non-negative advantage for the joint action, and it reduces the search space complexity from multiplicative to additive in the sizes of the action spaces.


\subsection{Multi-Agent Transformer (MAT)}

MAT is an encoder-decoder architecture that outputs agent actions from observations using the multi-agent advantage decomposition theorem, developed by \citet{wen2022multiagentreinforcementlearningsequence}.
It has theoretic monotonic improvement guarantees  and a search space that only increases linearly in the number of agents.
MAT's encoder maps agents' observation sequences to a latent representation used to estimate the value of the observations $V(\mathbf{o})$ and then fed into the decoder.
This is achieved using attention
\begin{equation}
    \text{Attention}(\mathbf{Q},\mathbf{K},\mathbf{V}) = \bm{\Lambda}\mathbf{V}, \qquad
    \bm{\Lambda} = \text{softmax}\left(\frac{\mathbf{Q}\mathbf{K}^T}{d_k}\right), \label{eqn:transformer_attn}
\end{equation}
where $\mathbf{Q},\mathbf{K}$ and $\mathbf{V}$ are the key, query and value matrices and $\bm{\Lambda}$ is called the attention matrix.

It employs self-attention blocks in both the encoder and decoder to filter observations and actions, respectively, where $\mathbf{Q},\mathbf{K}$ and $\mathbf{V}$ are derived from the relevant single input sequence.
An additional cross-attention block inside the decoder introduces information from agent observations during action-selection by making $\mathbf{Q}$ dependent on observations while $\mathbf{K}$ and $\mathbf{V}$ depend upon agent actions.

Note that action-selection differs between training and evaluation.
During evaluation, the decoder sequentially and auto-regressively transforms latent observation representations to actions, using masked (or causal) attention that prevents each agent viewing the `future' actions of teammates.
When masking, the attention matrix $\bm{\Lambda}$ becomes lower-triangular.
During training, action-selection can be carried out in parallel, sampling previously collected actions from the replay buffer.

At time of writing, MAT represents the state-of-the-art in MARL, surpassing the performance of HAPPO and MAPPO on many tasks.
MAT's use of attention allows it fine-grained control on the degree of importance each input token has on the output.
However, attention has its drawbacks.
MAT's attention mechanism has a time and memory cost that scales quadratically in the number of agents at inference.
Additionally, MAT cannot include any information outside its context window.
As a result, MAT will scale poorly as the number of agents becomes large.


\subsection{State-Space Models (SSMs)}

SSMs have recently gained popularity in deep learning applications.
SSMs are inspired by the classical state-space representation from control engineering, which uses a set of first-order differential equations to map an input to an output to describe the behaviour of a physical system.
The system's dynamics are determined by current observations and its existing state.
For a continuous and time-invariant system with an input signal $x(t) \in \mathbb{R}^D$, an SSM will output $y(t) \in \mathbb{R}^D$ via a hidden (or latent) state $h(t) \in \mathbb{R}^N$:
\begin{equation}
    \begin{aligned}
        h'(t) &= \mathbf{A}h(t) + \mathbf{B}x(t) \label{eqn:cont_SSM} \\
        y(t) &= \mathbf{C}h(t),
    \end{aligned}
\end{equation}
where $\mathbf{A} \in \mathbb{R}^{N \times N}$ is called the state matrix, $\mathbf{B} \in \mathbb{R}^{N \times D}$ is called the input matrix and $\mathbf{C} \in \mathbb{R}^{D \times N}$ is called the output matrix.
The matrices $\mathbf{A}, \mathbf{B}$ and $\mathbf{C}$ may all be made learnable during model training.

Often, we wish to evaluate discrete inputs $(x_0,...,x_t)$, transforming our continuous-time differential equation into a discrete-time difference equation.
Using a timestep size $\bm{\Delta}$, we can discretise \ref{eqn:cont_SSM} using various discretisation rules, replacing `continuous' parameters $\{\mathbf{A},\mathbf{B},\bm{\Delta}\}$ with `discrete' parameters $\{\mathbf{\bar{A}},\mathbf{\bar{B}}\}$.
With the zero-order hold discretisation method, we obtain
\begin{equation}
    \begin{aligned}
        \mathbf{\bar{A}} &= \exp(\bm{\Delta} \mathbf{A}) \label{eqn:disc_SSM_params} \\
        \mathbf{\bar{B}} &= \left( \bm{\Delta} \mathbf{A} \right)^{-1} \left( \exp \left(\bm{\Delta} \mathbf{A}\right) - \mathbf{I} \right) \cdot \bm{\Delta} \mathbf{B}.
    \end{aligned}
\end{equation}
Our SSM can now be evaluated for discrete inputs as a linear recurrence:
\begin{equation}
    \begin{aligned}
        h_t &= \mathbf{\bar{A}}h_{t-1} + \mathbf{\bar{B}}x_t \label{eqn:disc_SSM} \\
        y_t &= \mathbf{C}h_t.
    \end{aligned}
\end{equation}
In recurrent mode, an SSM is efficient when inferring (constant-time for a new sequence token) but slow during training (linear-time) since it is not parallelisable in this form.

The S4 \citep{gu2022efficientlymodelinglongsequences} massages this representation into an equivalent convolutional form under a linear time invariance assumption to allow parallel training, and the authors initialise $\mathbf{A}$ using the HiPPO initialisation scheme for further performance boosts \citep{gu2020hipporecurrentmemoryoptimal}.
Model design is a balance between efficiency and efficacy.
While the S4 is efficient, its dynamics are constant through time.
Therefore, it lacks the ability to adjust output in a content-aware manner, severely hampering effectiveness \citep{gu2024mambalineartimesequencemodeling}.

Mamba \citep{gu2024mambalineartimesequencemodeling} attempts to plug the gap left by S4; it is a different flavour of SSM that discards the parallelisable convolutional form in favour of content-awareness, termed a \textit{selective} SSM.
Mamba allows some of its parameters--namely $\mathbf{B}, \mathbf{C}$ and $\bm{\Delta}$--to depend on its inputs $x_t$, in particular
\begin{equation}
    \begin{aligned}
        \mathbf{B}_t &= \text{Linear}_\mathbf{B}(x_t) \label{eqn:selective_SSM_params} \\
        \mathbf{C}_t &= \text{Linear}_\mathbf{C}(x_t) \\
        \bm{\Delta}_t &= \text{softplus} \left( \text{Linear}_{\bm{\Delta}} (x_t) \right),
    \end{aligned}  
\end{equation}
where $\text{softplus}(\cdot) = \log \left( 1 + \exp(\cdot) \right)$ and $\text{Linear}(\cdot)$ is a linear projection.
Additionally, \ref{eqn:disc_SSM} can be written as an associative scan \citep{martin2018parallelizinglinearrecurrentneural}, which allows us to use an efficient parallel algorithm \citep{blelloch1990prefixsumsandtheir} to offset the higher time-cost associated with training purely in recurrent mode.
Mamba's expanded expressivity and computational efficiency allows it to deliver high-quality performance to match the Transformer, particularly for long sequences.

A vanilla Mamba $\textit{module}$ is depicted in the leftmost schematic in Figure \ref{fig:mamba_and_bimamba}.
The input sequence is duplicated and expanded to twice the input dimension.

One duplicate is processed through a causal convolution, then passed through the SiLU/Switch activation function \citep{ramachandran2017searchingactivationfunctions}, and finally sent through the selective SSM.
The other duplicate undergoes a SiLU activation and acts as a gate for the SSM output.
The gated output is then projected back down to the original input dimension.
The Mamba module in Figure \ref{fig:mamba_and_bimamba} is wrapped with an additional normalisation layer and residual connection, forming what we will term a Mamba $\textit{block}$ in this paper.
Many Mamba blocks can be chained together in this fashion, using an outermost normalisation layer for the final output; the authors use LayerNorm \citep{ba2016usingfastweightsattend}.



\section{Multi-Agent Mamba (MAM)}

In this section, we introduce MAM as a scalable alternative to MAT, which replaces MAT's attention with Mamba blocks.


\subsection{Encoder}


\begin{figure}
    \centering
    \begin{minipage}[b]{0.48\linewidth}
        \centering
        \includegraphics[width=\linewidth]{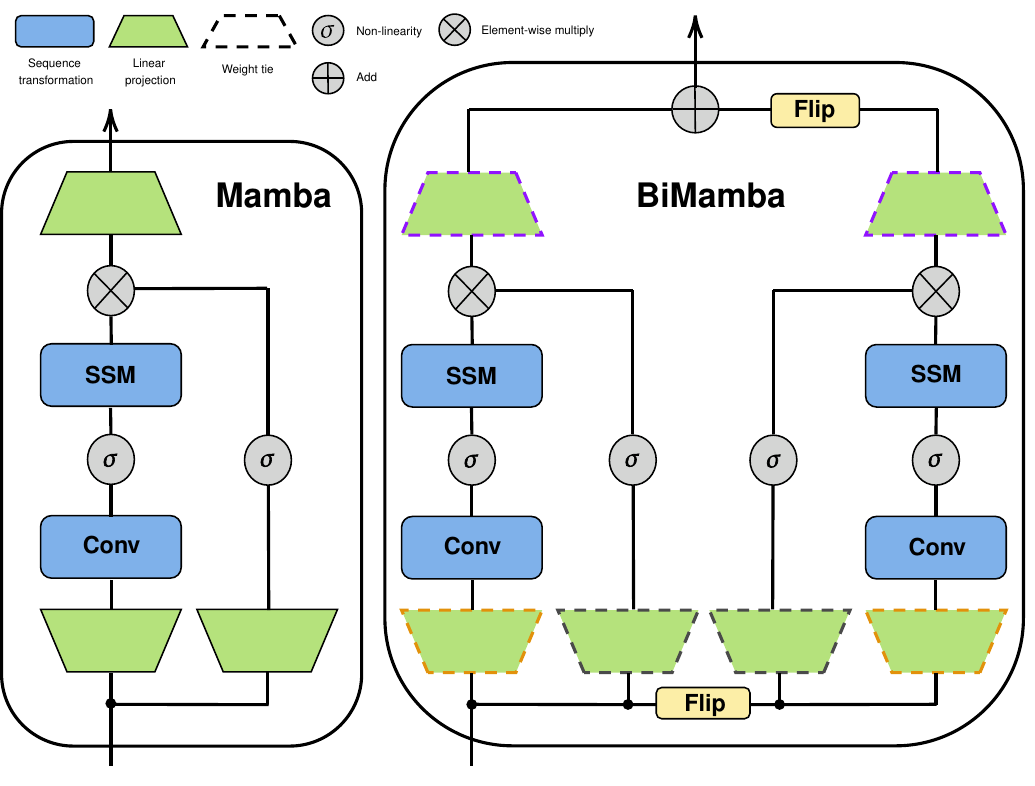}
        \vspace{0pt}
        \caption{The vanilla and bi-directional Mamba modules used to replace causal and non-causal attention, respectively.
Left: The original left-to-right causal Mamba module proposed by \citet{gu2024mambalineartimesequencemodeling}.
Right: A bi-directional extension of the Mamba module proposed by \citet{schiff2024caduceusbidirectionalequivariantlongrange}.}
        \label{fig:mamba_and_bimamba}
    \end{minipage}
    \hfill
    \begin{minipage}[b]{0.48\linewidth}
        \centering
        \includegraphics[width=\linewidth]{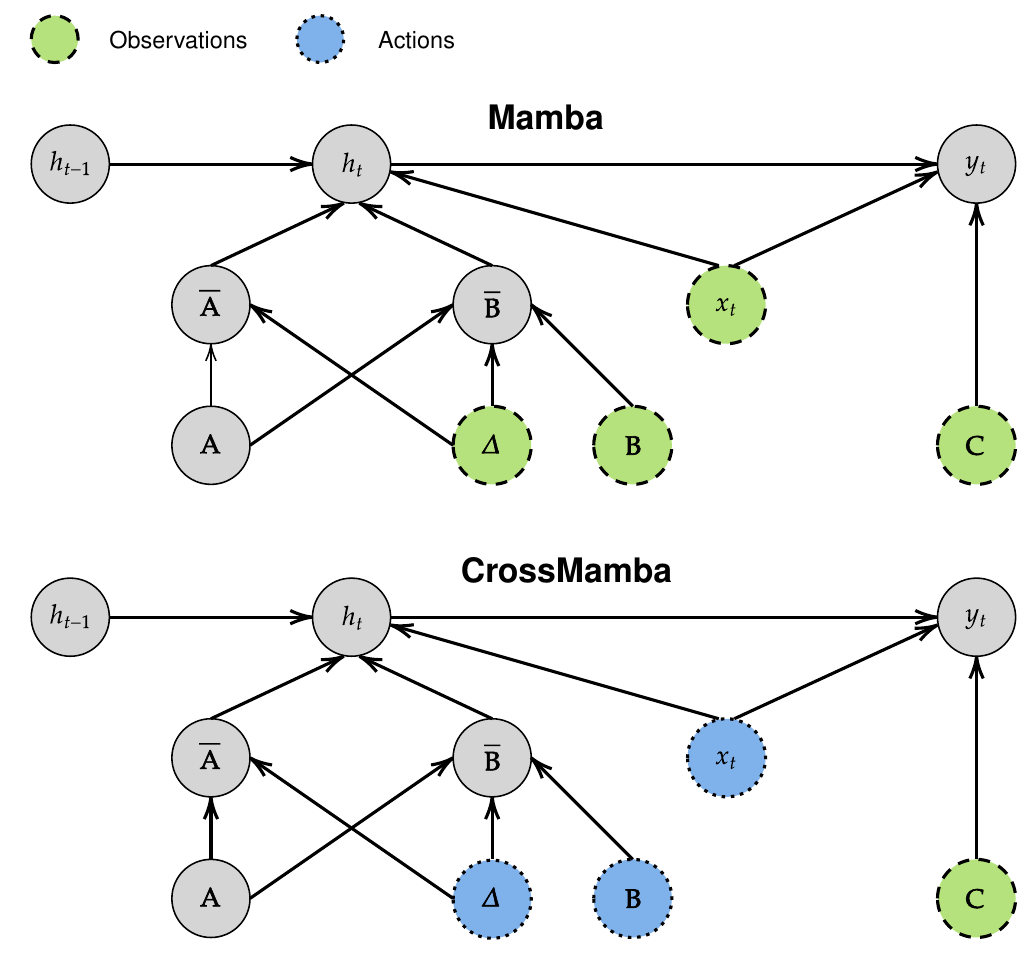}
        \caption{The dependency chart of the SSM layer for a vanilla Mamba module (top) and a CrossMamba module (bottom) in a single timestep $t$.
We use colour to highlight which input sequence each block's selective parameters depend upon.}
        \label{fig:mamba_and_crossmamba}
    \end{minipage}
\end{figure}


A vanilla Mamba block is causal, processing a single input sequence from left to right \citep{gu2024mambalineartimesequencemodeling}.
However, self-attention links tokens in either direction; in particular, MAT uses unmasked self-attention in its encoder \citep{wen2022multiagentreinforcementlearningsequence}.
Therefore, we adopt a modified, bi-directional Mamba block that allows observation representations to encode information from every agent's local view, as is allowed in MAT \citep{wen2022multiagentreinforcementlearningsequence}.
We use the architecture laid out by \citet{schiff2024caduceusbidirectionalequivariantlongrange}, shown in the rightmost schematic in Figure \ref{fig:mamba_and_bimamba}.
While originally applied to DNA sequence modelling, its architecture is domain-agnostic.
The bi-directional Mamba block applies the original Mamba module twice.
We begin by applying the Mamba module to the original input sequence and then to a reversed copy along the length (or agent) dimension.
To merge the results, we flip the reversed output along its length and add it to the forward output.

To avoid doubling the number of parameters in a bi-directional Mamba block compared to vanilla Mamba, we share parameters.
The `forward' and `reverse' Mamba modules, which account for a large portion of the block's parameters \citep{gu2024mambalineartimesequencemodeling}, instead use the same projection weights.


\subsection{Decoder}

In the decoder, it is important to ensure the current agent only accesses the information of preceding agents.
MAT achieves this using causal masking \citep{ba2016usingfastweightsattend}, but the Mamba module is causal out-of-the-box.
This makes causal self-attention replacement straightforward: simply use Mamba as a drop-in replacement, feeding agent actions through the vanilla Mamba block to obtain latent action representations.

However, replacement of MAT's cross-attention requires more care.
Vanilla Mamba only processes a single input sequence, but cross-attention attends between two different sequences--latent observation and action representations in MAT's case.
We adapt Mamba to process information from two inputs, creating a `cross-attentional' Mamba block which we name CrossMamba.
Following the work of \citet{ali2024hiddenattentionmambamodels}, unrolling \ref{eqn:disc_SSM} and using the initial condition $h_0=0$ gives
\begin{equation}
    \begin{aligned}
        h_t &= \sum^t_{j=1} \left( \prod^t_{k=j+1} \mathbf{\bar{A}}_k \right) \mathbf{\bar{B}}_j x_j \label{eqn:unrolled_mamba_attn} \\
        y_t &= \mathbf{C}_t \sum^{t}_{j=1} \left( \prod^t_{k=j+1} \mathbf{\bar{A}}_k \right) \mathbf{\bar{B}}_j x_j,
    \end{aligned}
\end{equation}
which can be converted into the following matrix form:
\begin{equation}
    \begin{aligned}
        \mathbf{y} &= \bm{\tilde{\Lambda}}\mathbf{x} \\
    \begin{bmatrix}
        y_1 \\
        y_2 \\
        \vdots \\
        y_L
    \end{bmatrix}
    &=
    \begin{bmatrix}
        \mathbf{C}_1\mathbf{\bar{B}}_1 & 0 & \hdots & 0 \\
        \mathbf{C}_2\mathbf{\bar{A}}_2\mathbf{\bar{B}}_2 & \mathbf{C}_2\mathbf{\bar{B}}_2 & \hdots & 0 \\
        \vdots & \vdots & \ddots & \vdots \\
        \mathbf{C}_L\prod^L_{k=2}\mathbf{\bar{A}}_k\mathbf{\bar{B}}_1 & \mathbf{C}_L\prod^{L}_{k=3}\mathbf{\bar{A}}_k\mathbf{\bar{B}}_2 & \hdots & \mathbf{C}_L\mathbf{\bar{B}}_L
    \end{bmatrix}
    \begin{bmatrix}
        x_1 \\
        x_2 \\
        \vdots \\
        x_L
    \end{bmatrix}, \label{eqn:mamba_attn_matrix}
    \end{aligned}
\end{equation}
where $L$ is the sequence length.
For a single element of $\bm{\tilde{\Lambda}}$, we have in general
\begin{equation}
    \tilde{\Lambda}_{ij} = \mathbf{C}_i \left( \prod^i_{k=j+1}\mathbf{\bar{A}}_k \right) \mathbf{\bar{B}}_j.
\label{eqn:mamba_attn_element}
\end{equation}
Given \ref{eqn:disc_SSM_params} and \ref{eqn:selective_SSM_params}, we can rewrite \ref{eqn:mamba_attn_element} as
\begin{equation}
    \begin{aligned}
        \tilde{\Lambda}_{ij} &= S_\mathbf{C}(x_i) 
        \left( \prod_{k=j+1}^{i} \exp \left( \text{softplus}(S_{\bm{\Delta}}(x_k)) \mathbf{A} \right) \right) \cdot \\
        & \quad \Biggl( \left( \text{softplus} \left( S_{\bm{\Delta}}(x_j) \right) \mathbf{A} \right)^{-1} \left( \exp \left( \text{softplus}(S_{\bm{\Delta}}(x_j))A \right) - \mathbf{I} \right) \cdot \\
        & \quad \text{softplus} \left( S_{\bm{\Delta}}(x_j) \right) S_\mathbf{B}(x_j) \Biggr) \\
        &= \mathbf{\tilde{Q}}_i \mathbf{H}_{i,j} \mathbf{\tilde{K}}_j, \label{eqn:mamba_attn_element_QHK}
    \end{aligned}
\end{equation}
where $\mathbf{\tilde{Q}}$ and $\mathbf{\tilde{K}}$ are query and key matrices, respectively, and $\mathbf{H}$ is an additional matrix unique to Mamba that summarises the historical context between sequence tokens $x_j$ to $x_i$.
If we let $\mathbf{x}=\mathbf{\tilde{V}}$, then the reformulations in \ref{eqn:mamba_attn_matrix} and \ref{eqn:mamba_attn_element_QHK} show that a vanilla Mamba block possesses a form of causal self-attention similar to \ref{eqn:transformer_attn} \citep{ali2024hiddenattentionmambamodels}.
CrossMamba, in contrast to a vanilla Mamba block, allows the selective parameter $\mathbf{C}$ to depend on a second input sequence.

In our MARL cross-attentional case, agent observations form the target sequence and agent observations make up the source sequence which we wish to `attend' to during auto-regressive action-selection.
This is highlighted in Figure \ref{fig:mamba_and_crossmamba}.



\section{Experiments}


\begin{figure}
    \centering
    \includegraphics[width=0.32\linewidth]{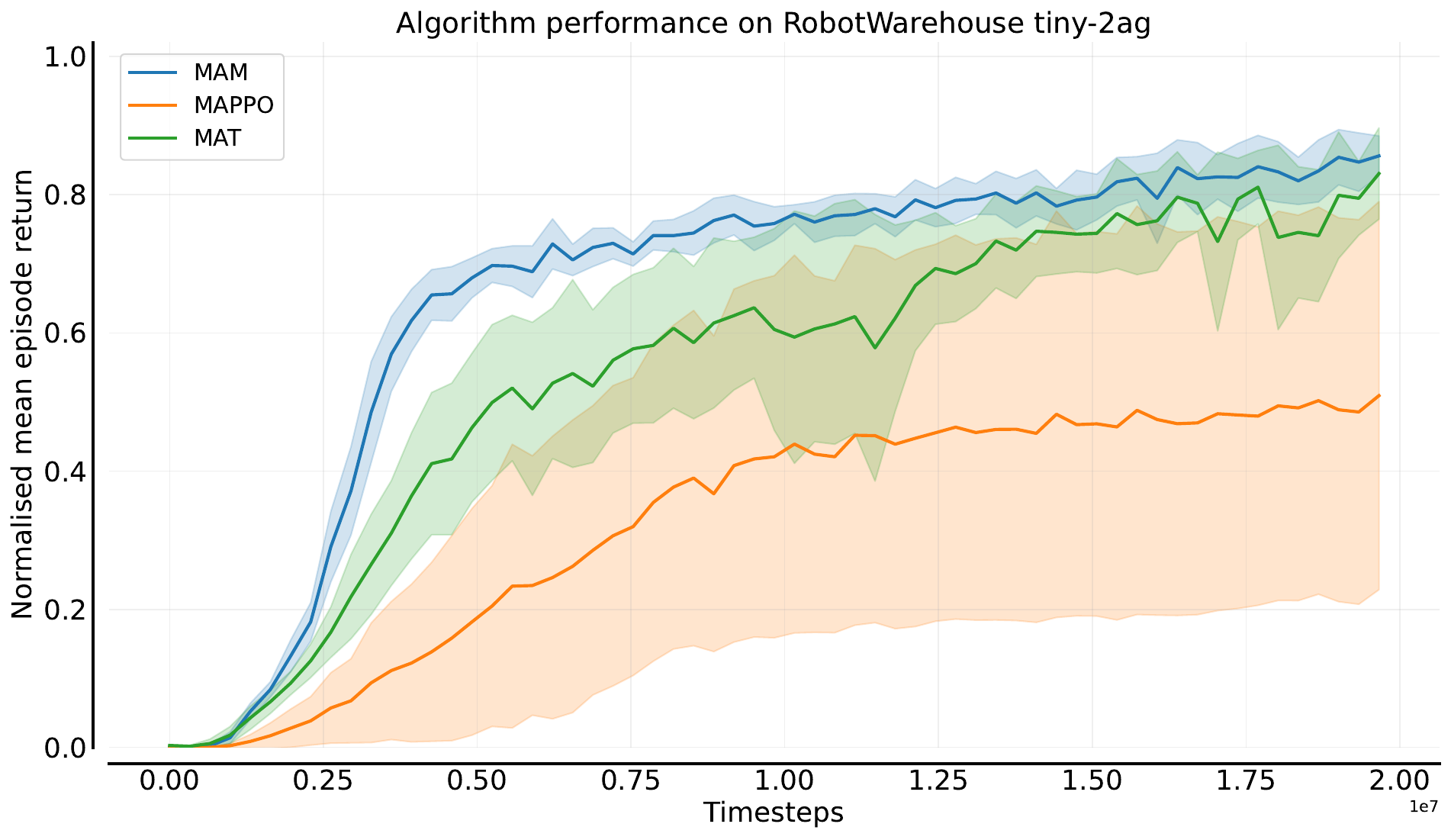}
    \includegraphics[width=0.32\linewidth]{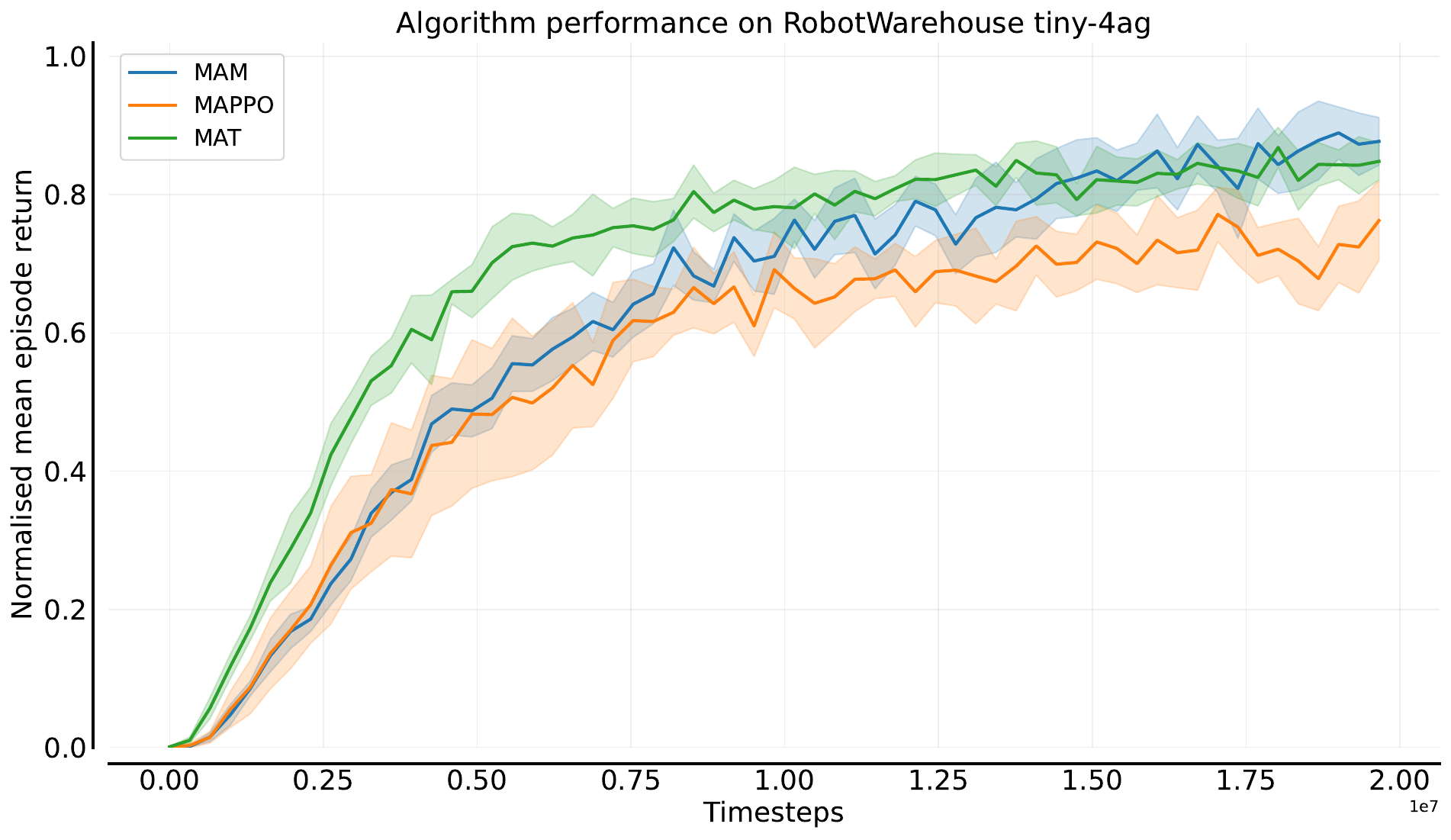}
    \includegraphics[width=0.32\linewidth]{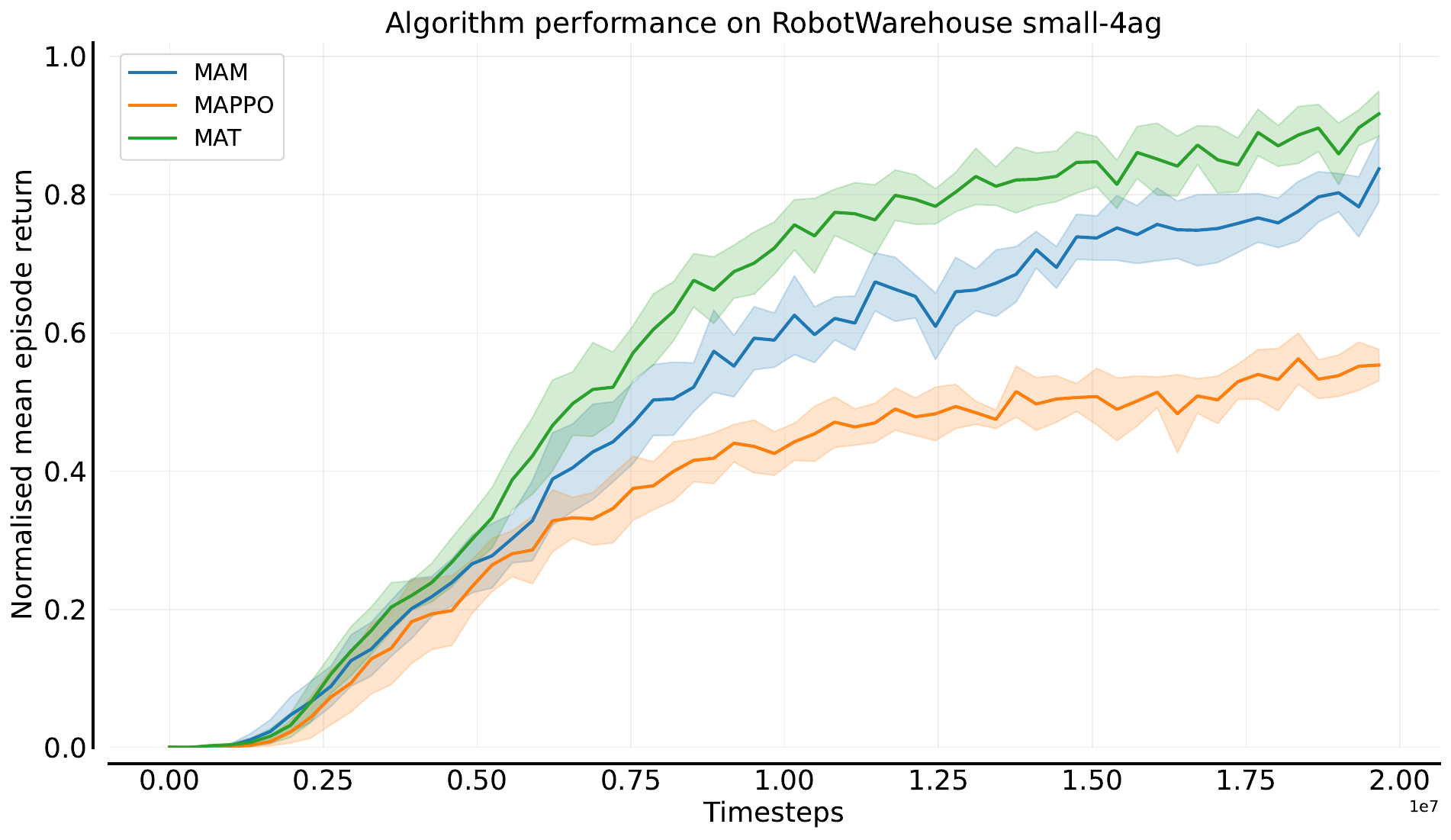}
    \includegraphics[width=0.32\linewidth]{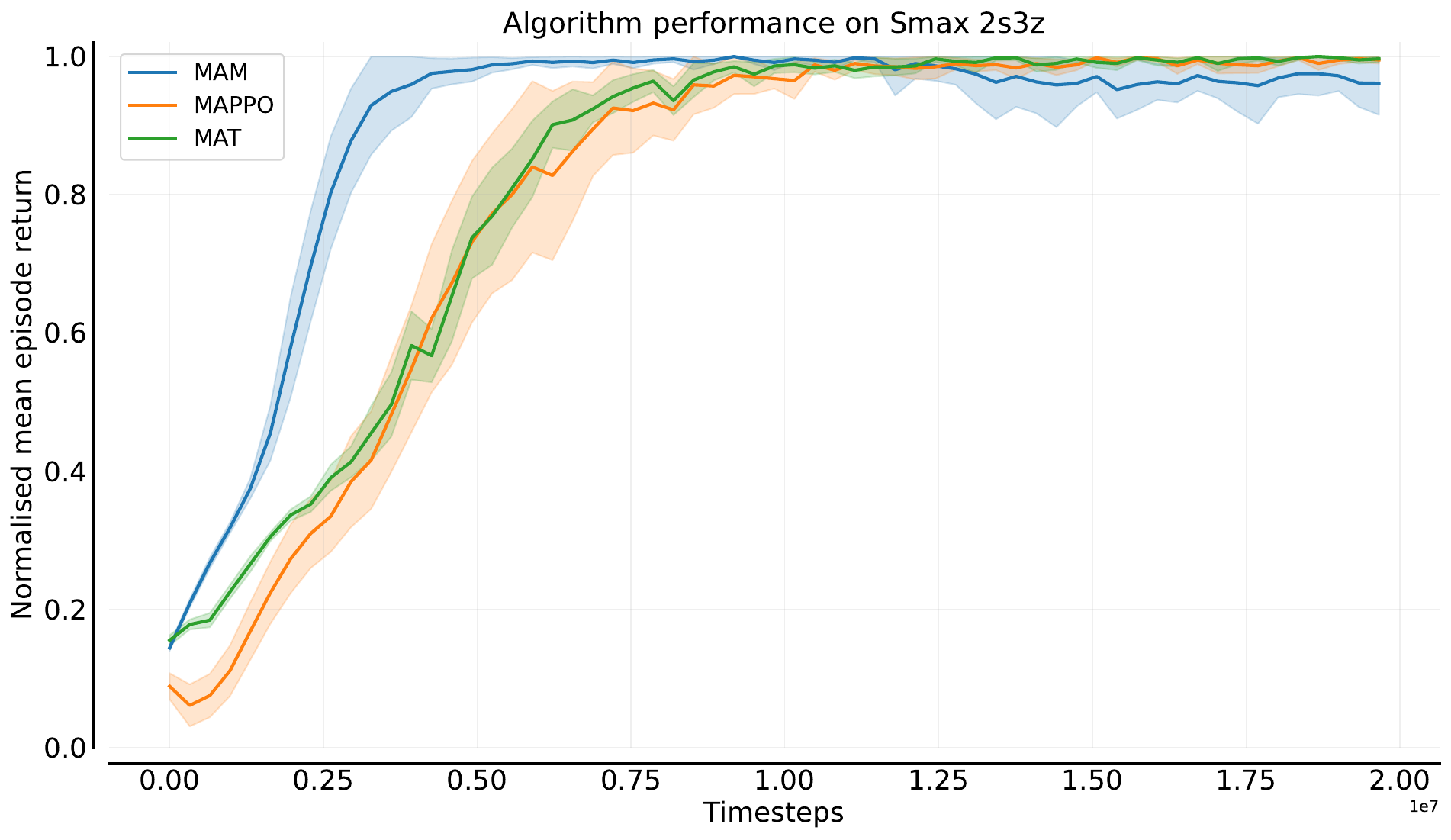}
    \includegraphics[width=0.32\linewidth]{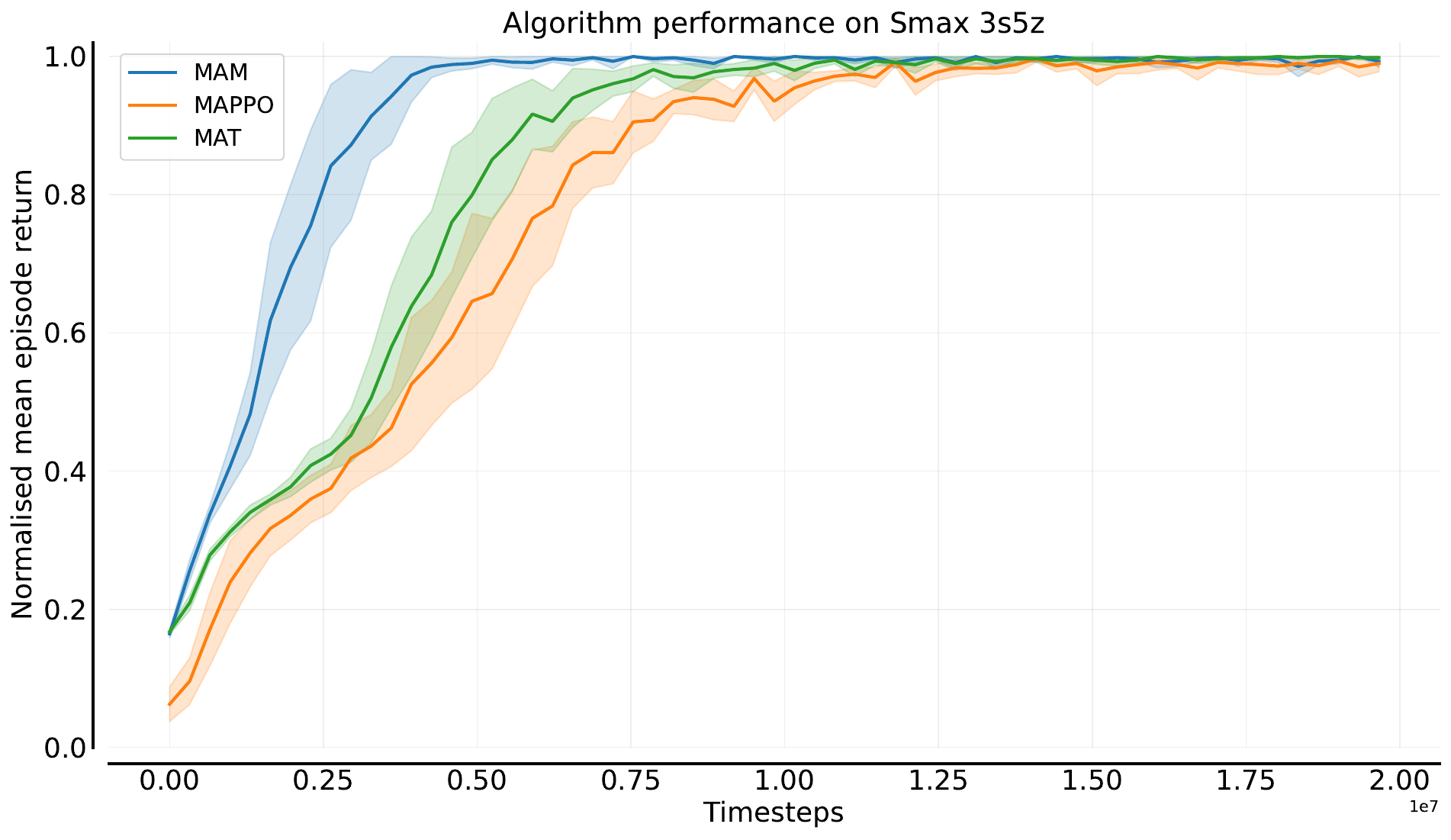}
    \includegraphics[width=0.32\linewidth]{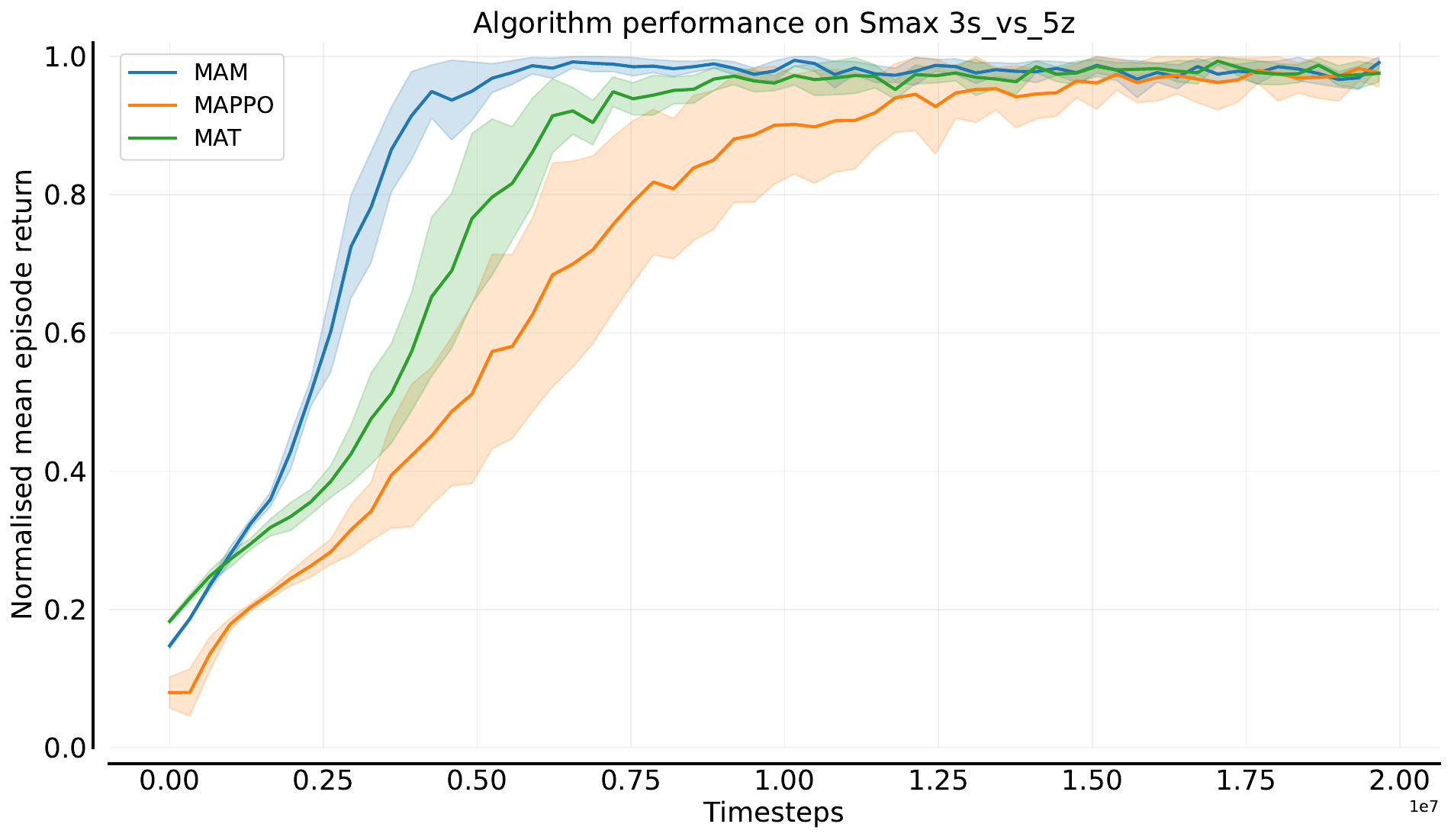}
    \includegraphics[width=0.32\linewidth]{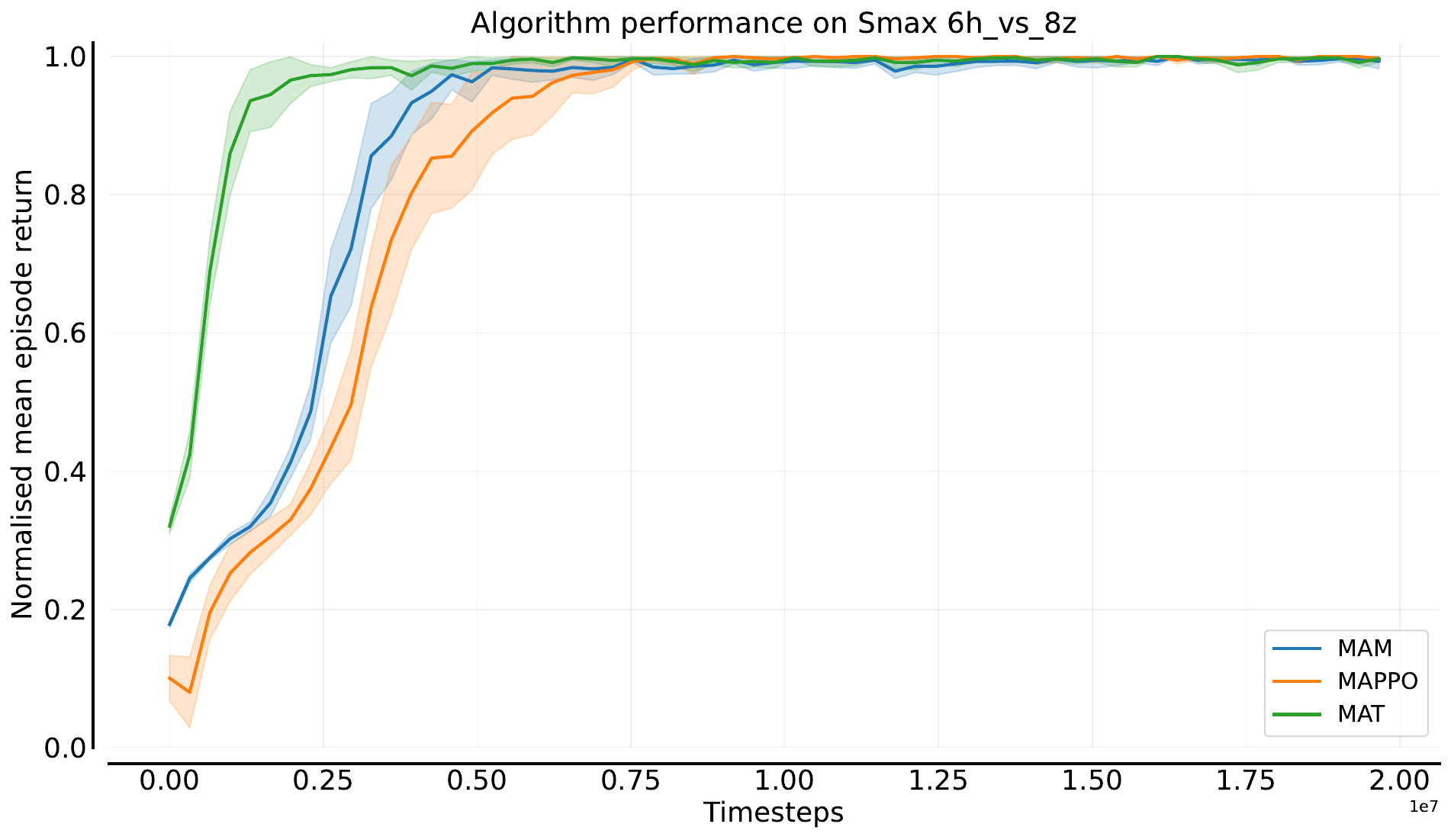}
    \includegraphics[width=0.32\linewidth]{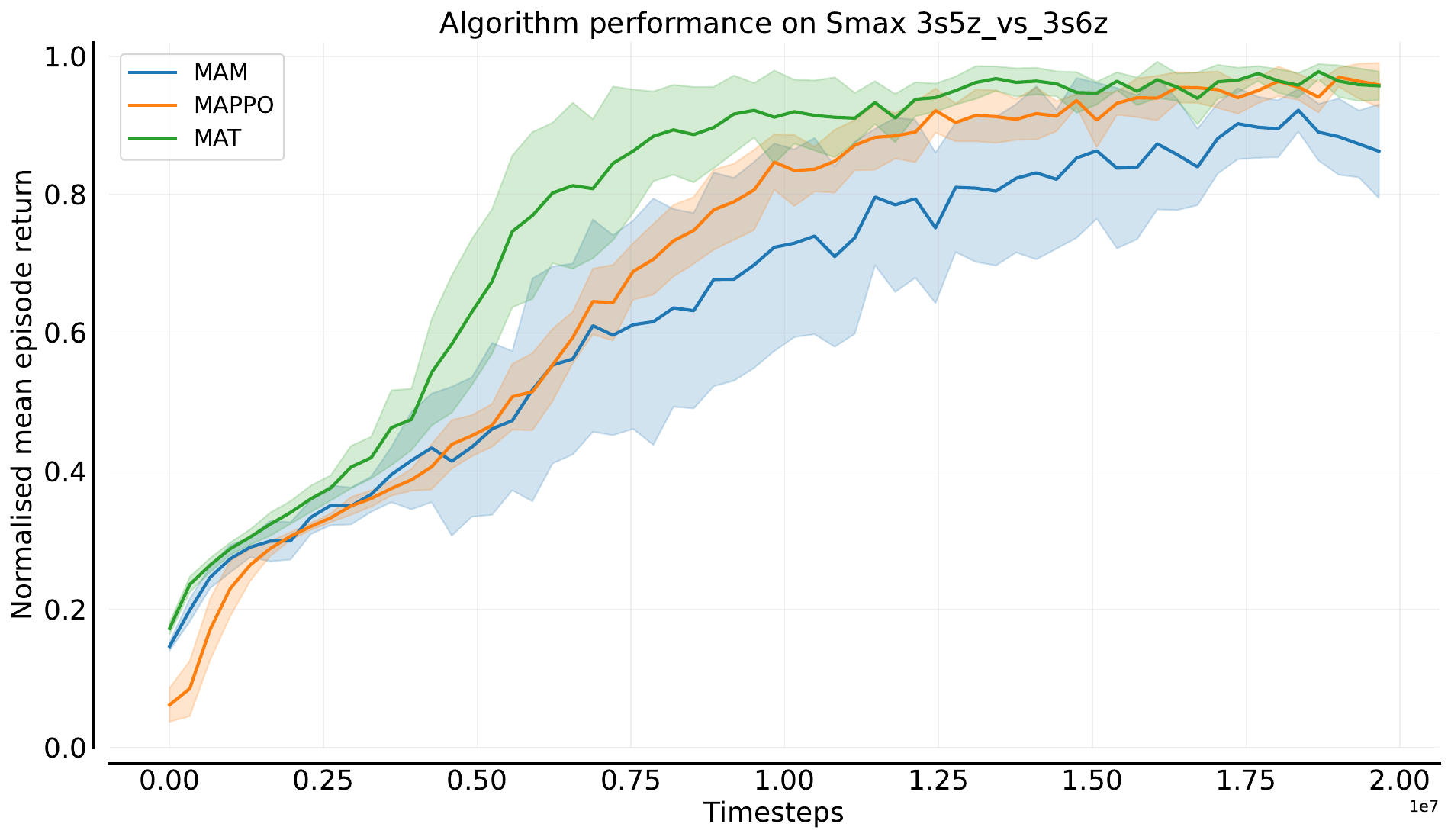}
    \includegraphics[width=0.32\linewidth]{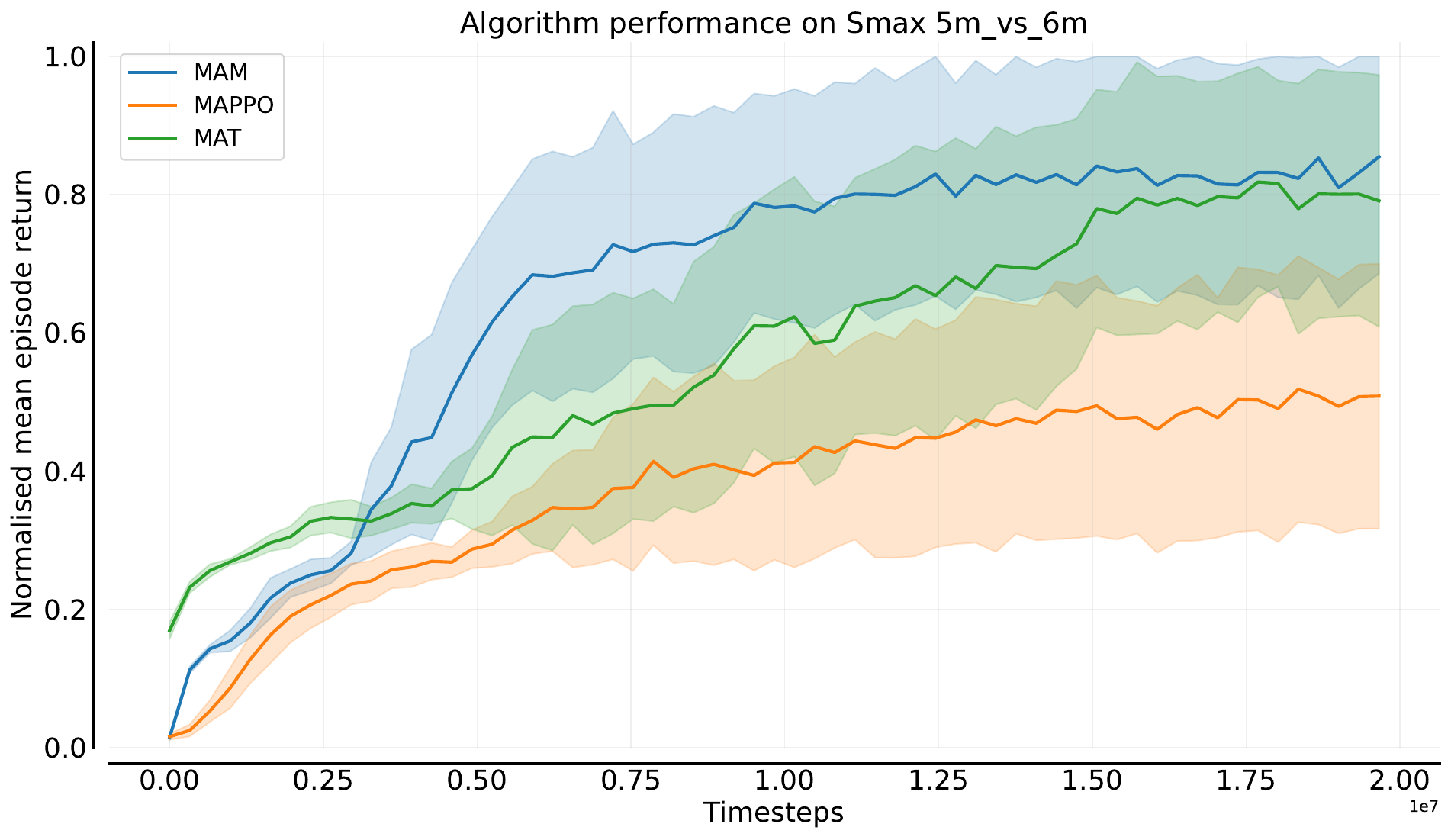}
    \includegraphics[width=0.32\linewidth]{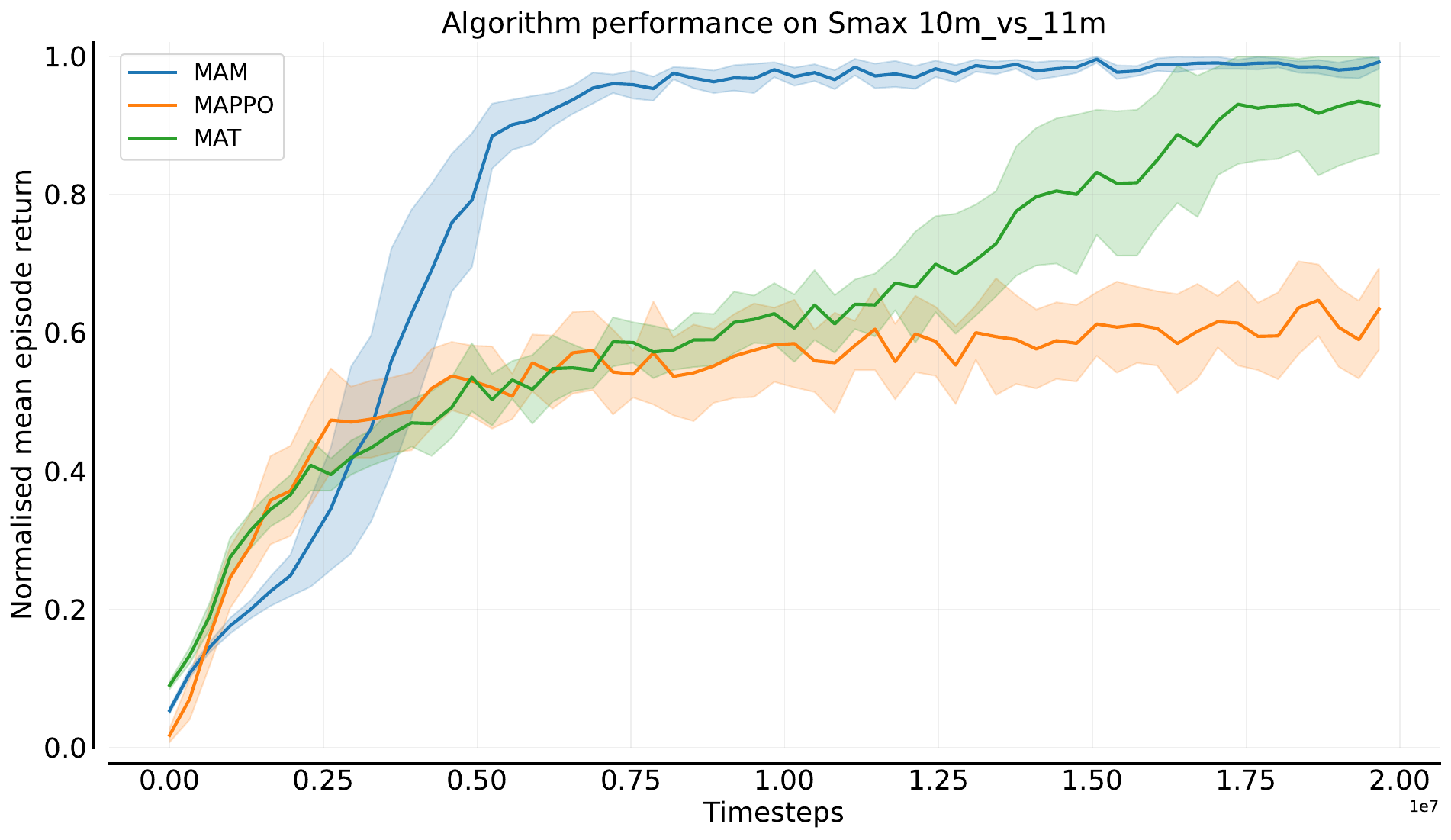}
    \includegraphics[width=0.32\linewidth]{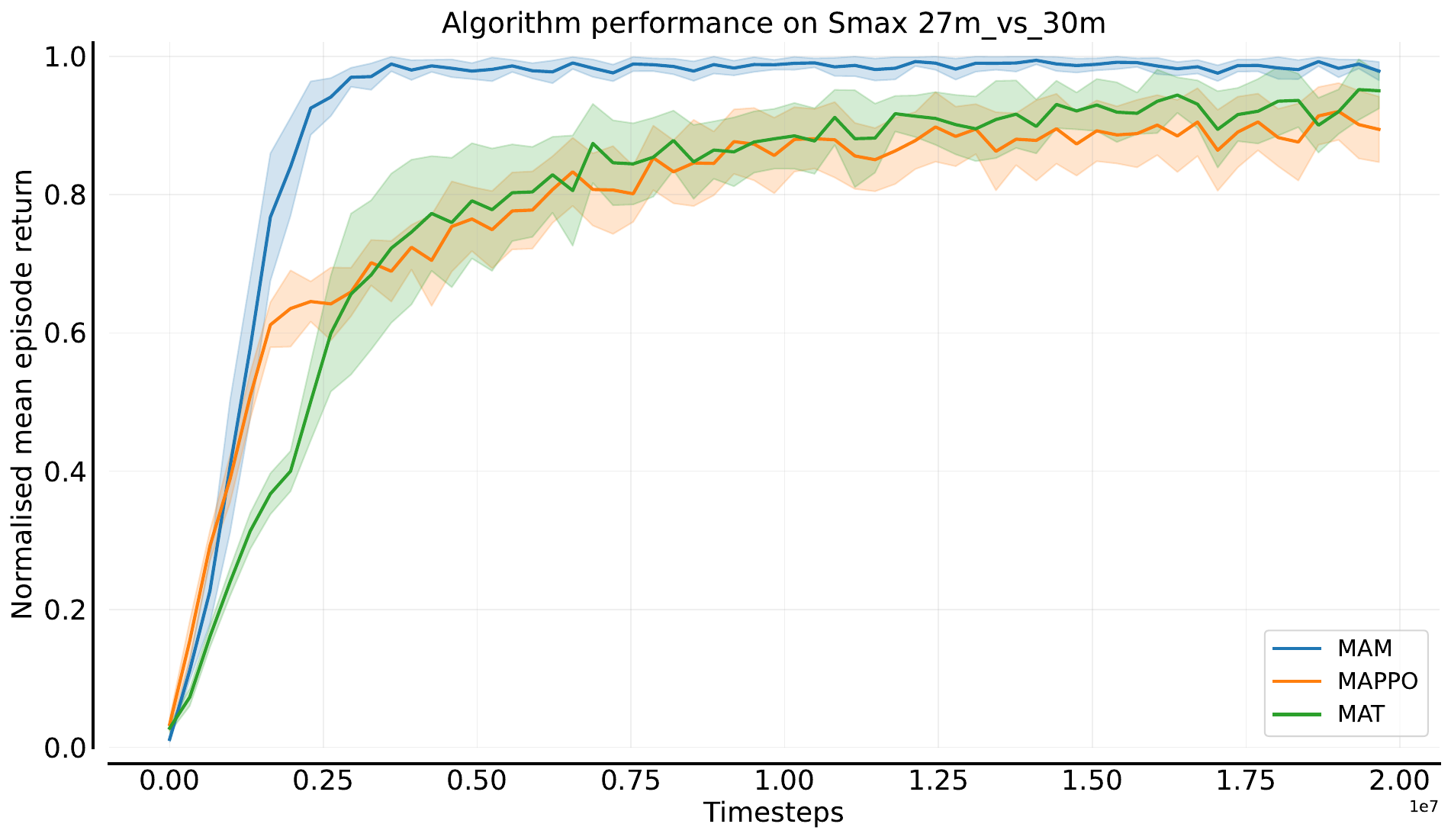}
    \includegraphics[width=0.32\linewidth]{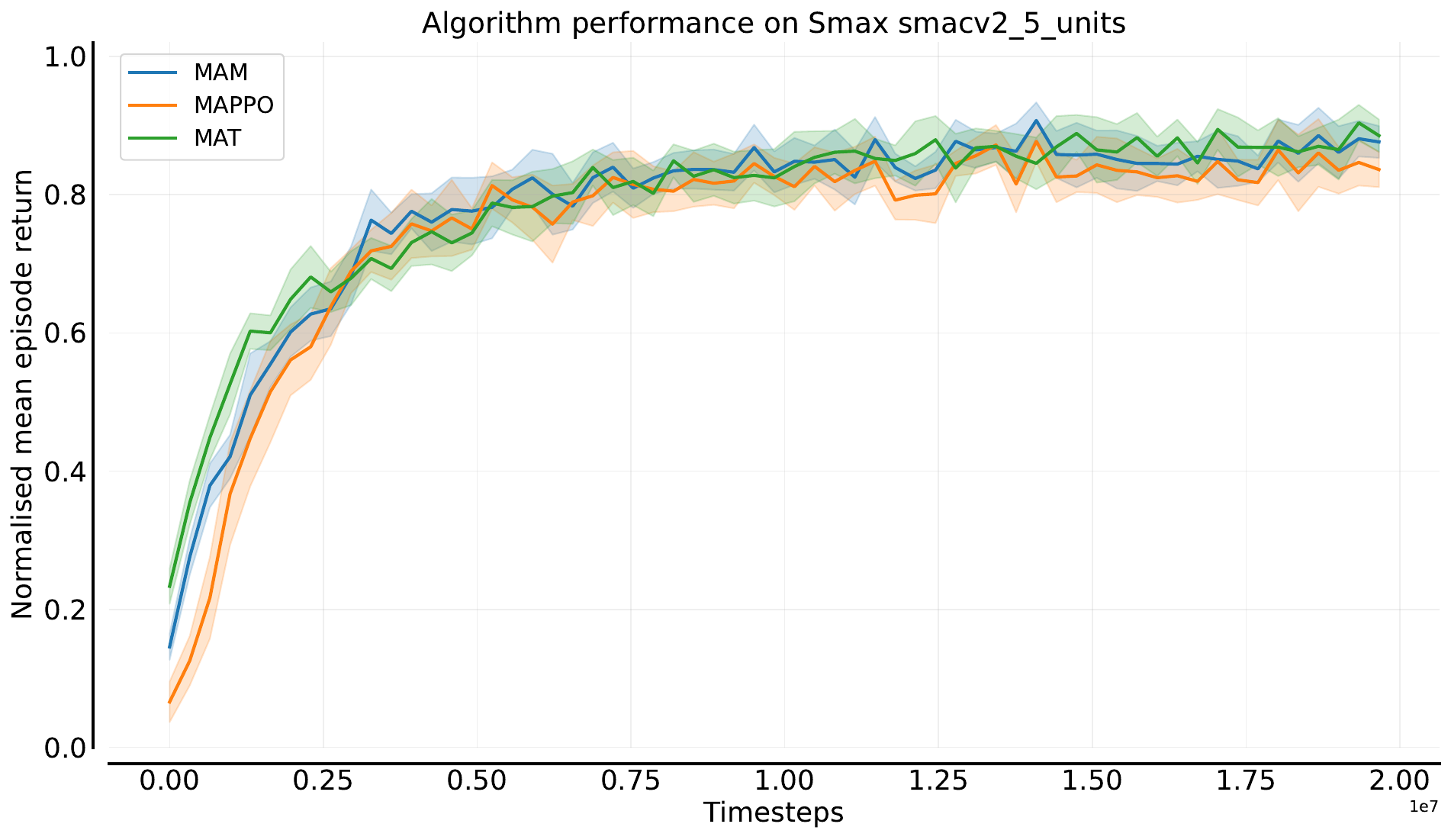}
    \includegraphics[width=0.32\linewidth]{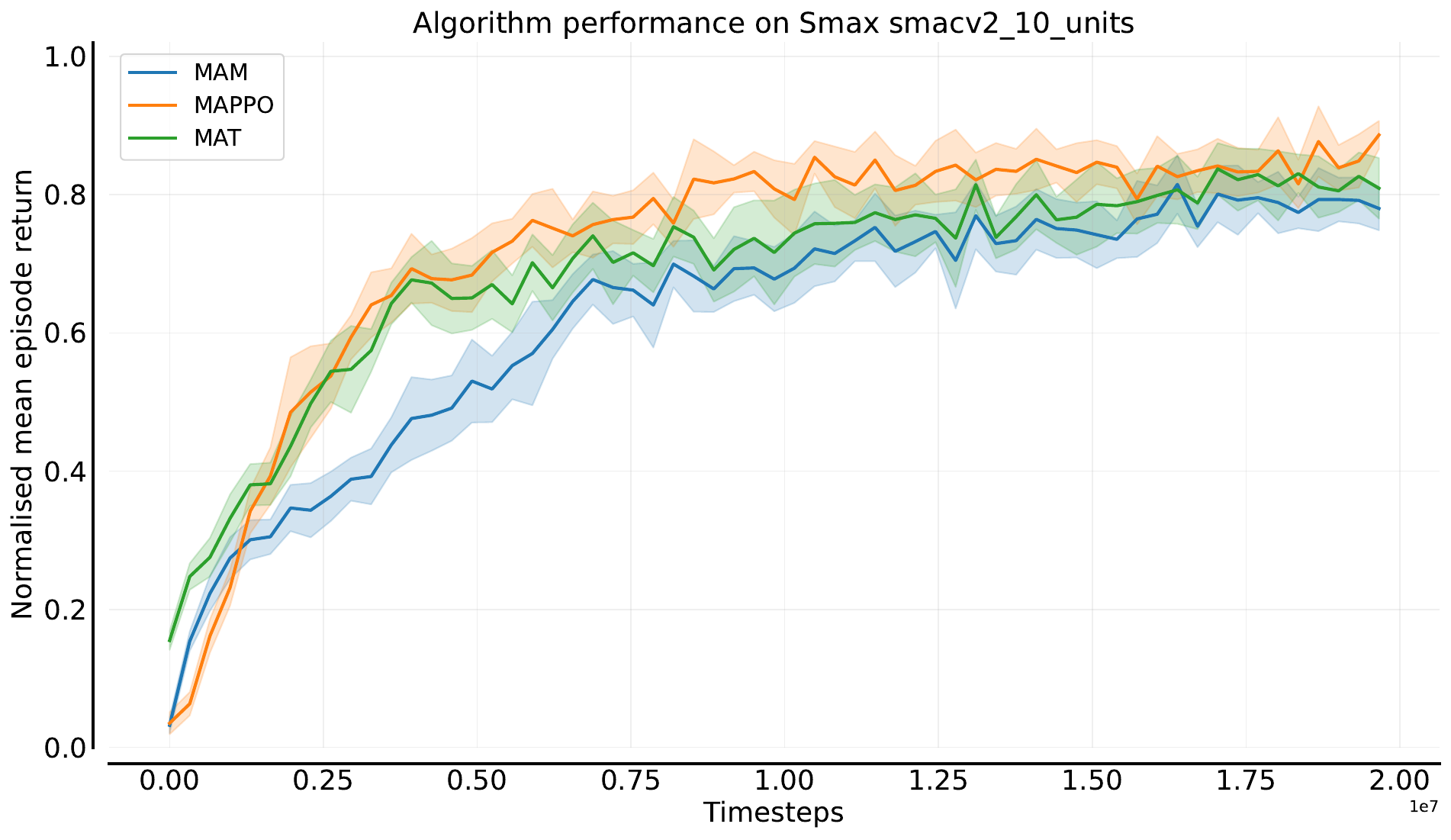}
    \includegraphics[width=0.32\linewidth]{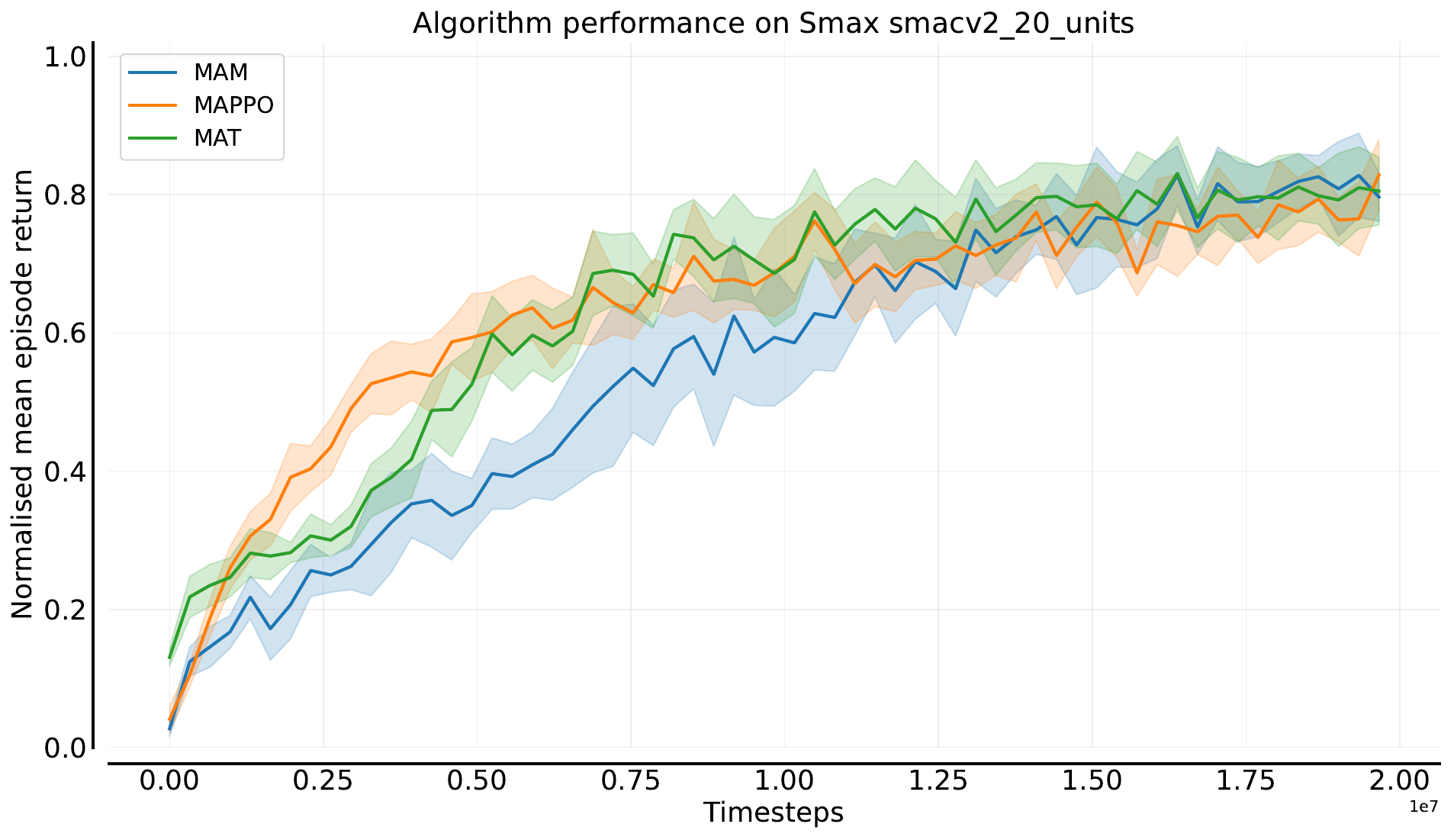}
    \includegraphics[width=0.32\linewidth]{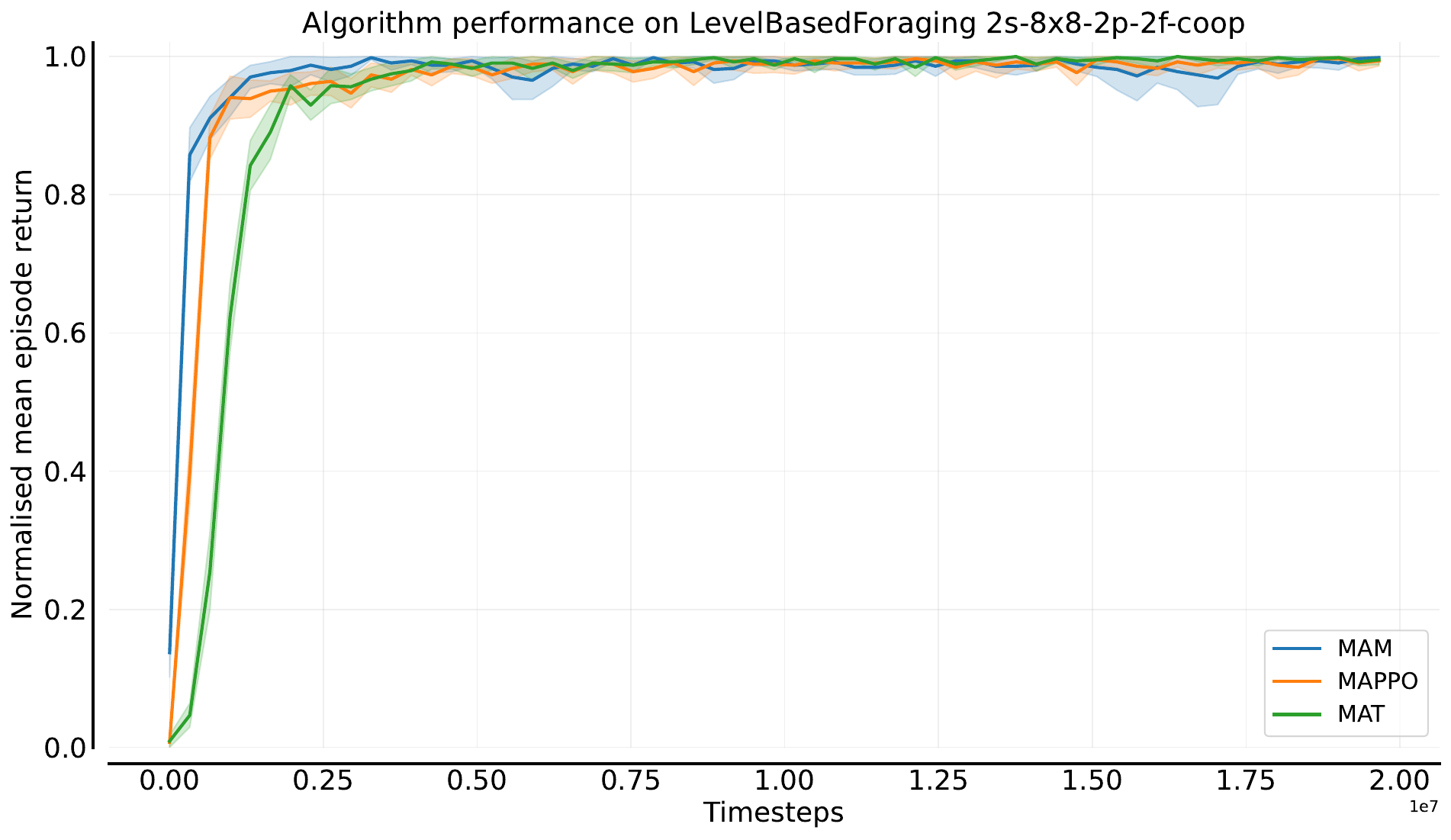}
    \includegraphics[width=0.32\linewidth]{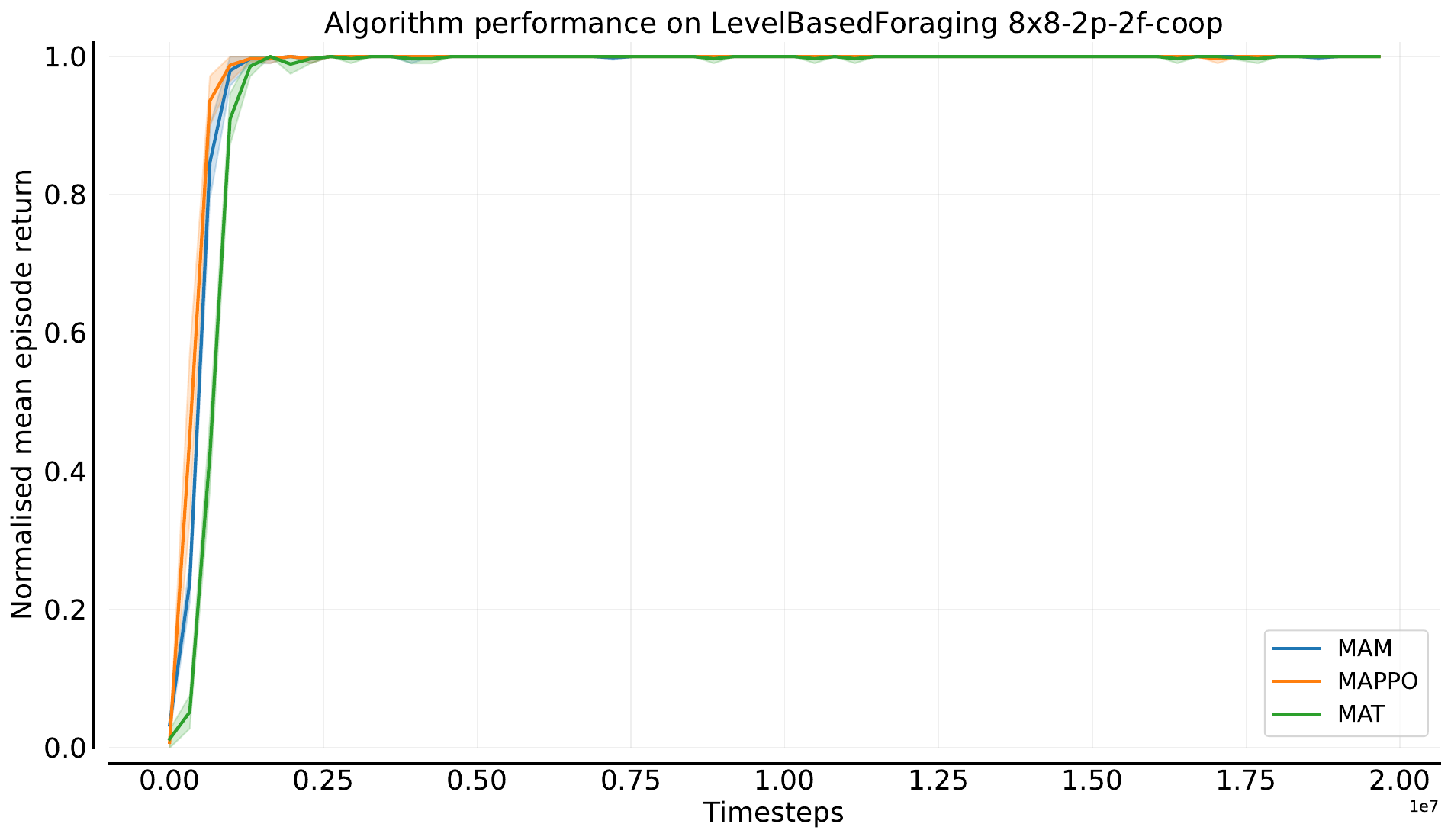}
    \includegraphics[width=0.32\linewidth]{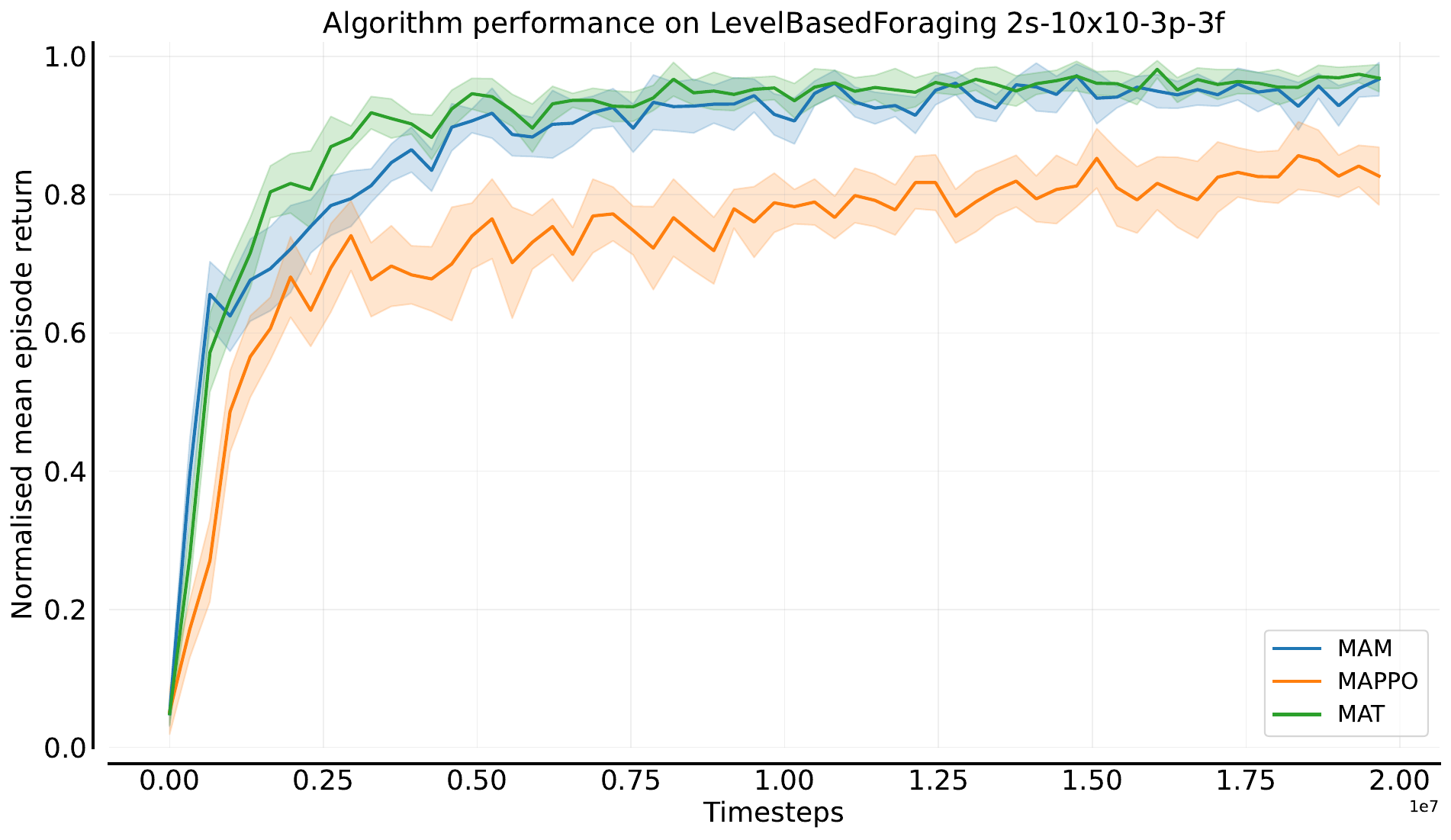}
    \includegraphics[width=0.32\linewidth]{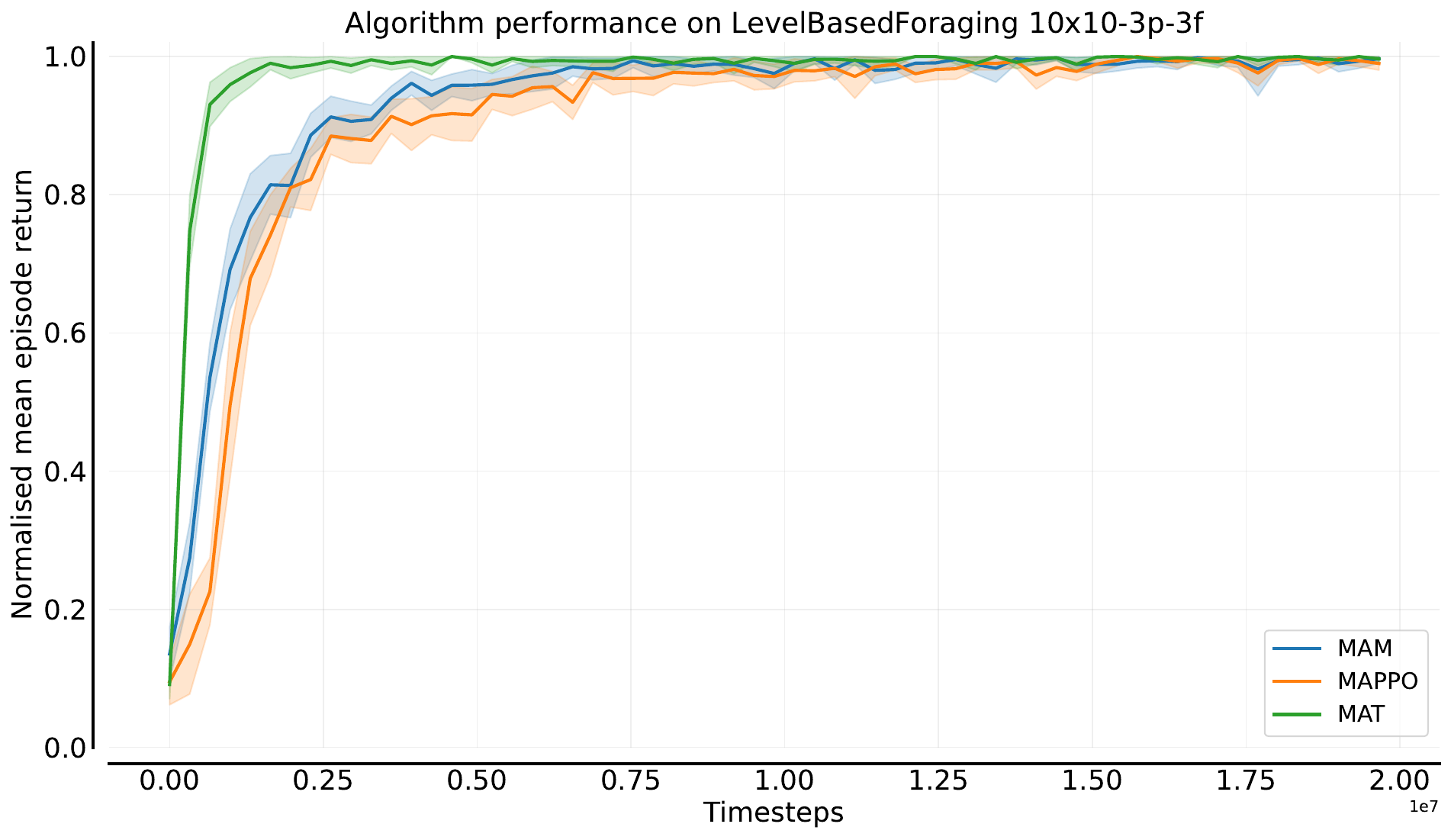}
    \includegraphics[width=0.32\linewidth]{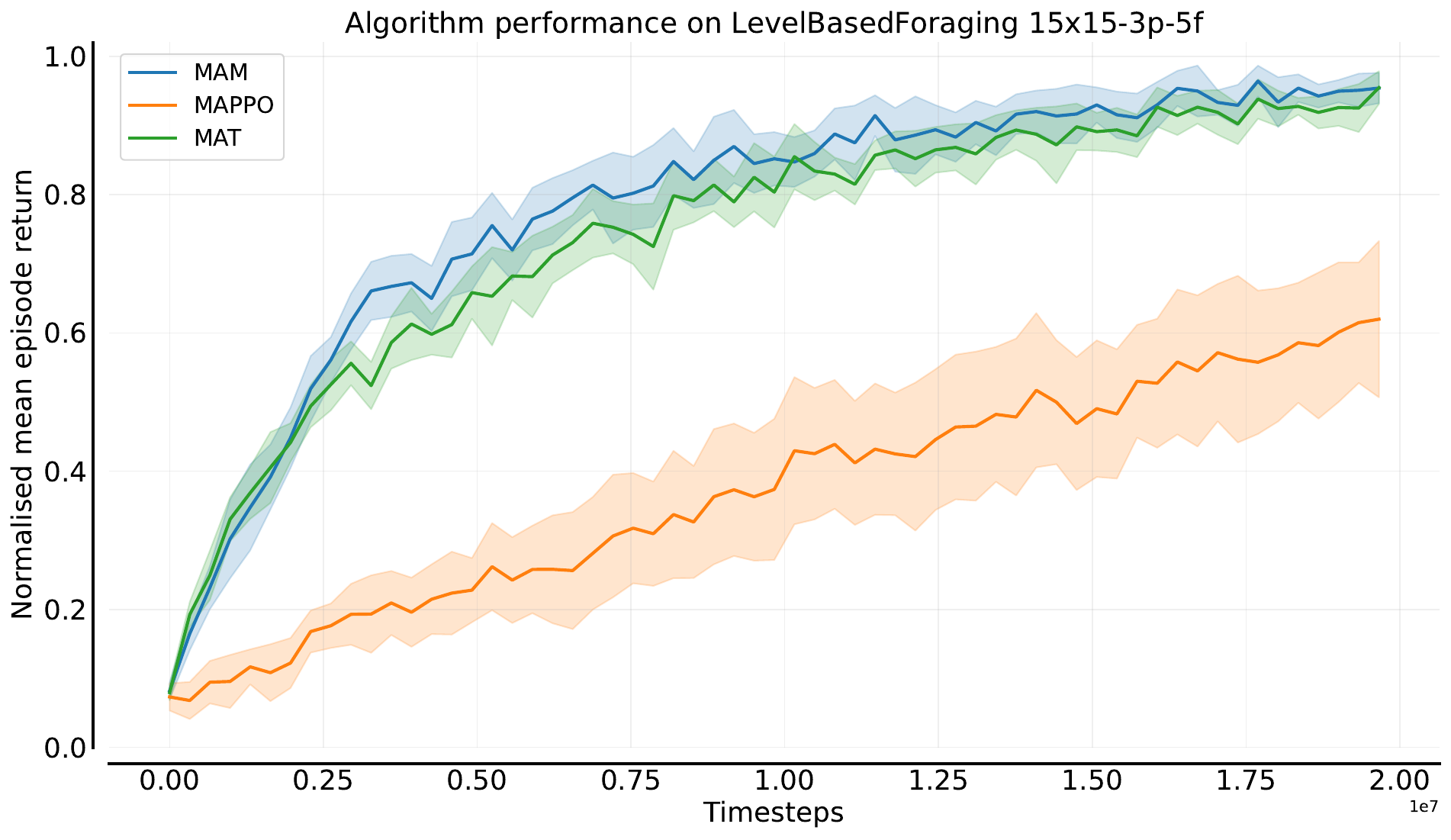}
    \includegraphics[width=0.32\linewidth]{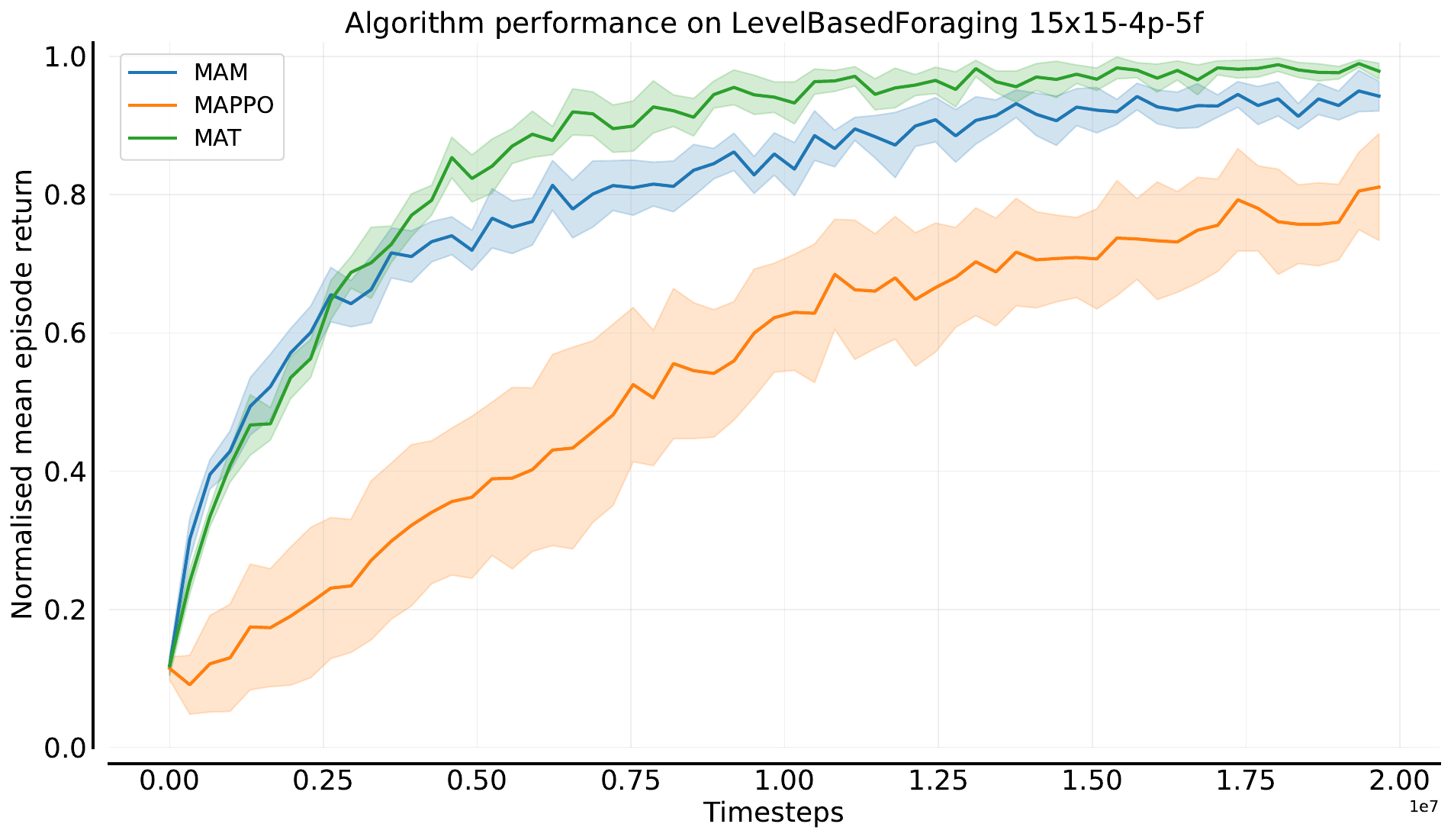}
    \includegraphics[width=0.32\linewidth]{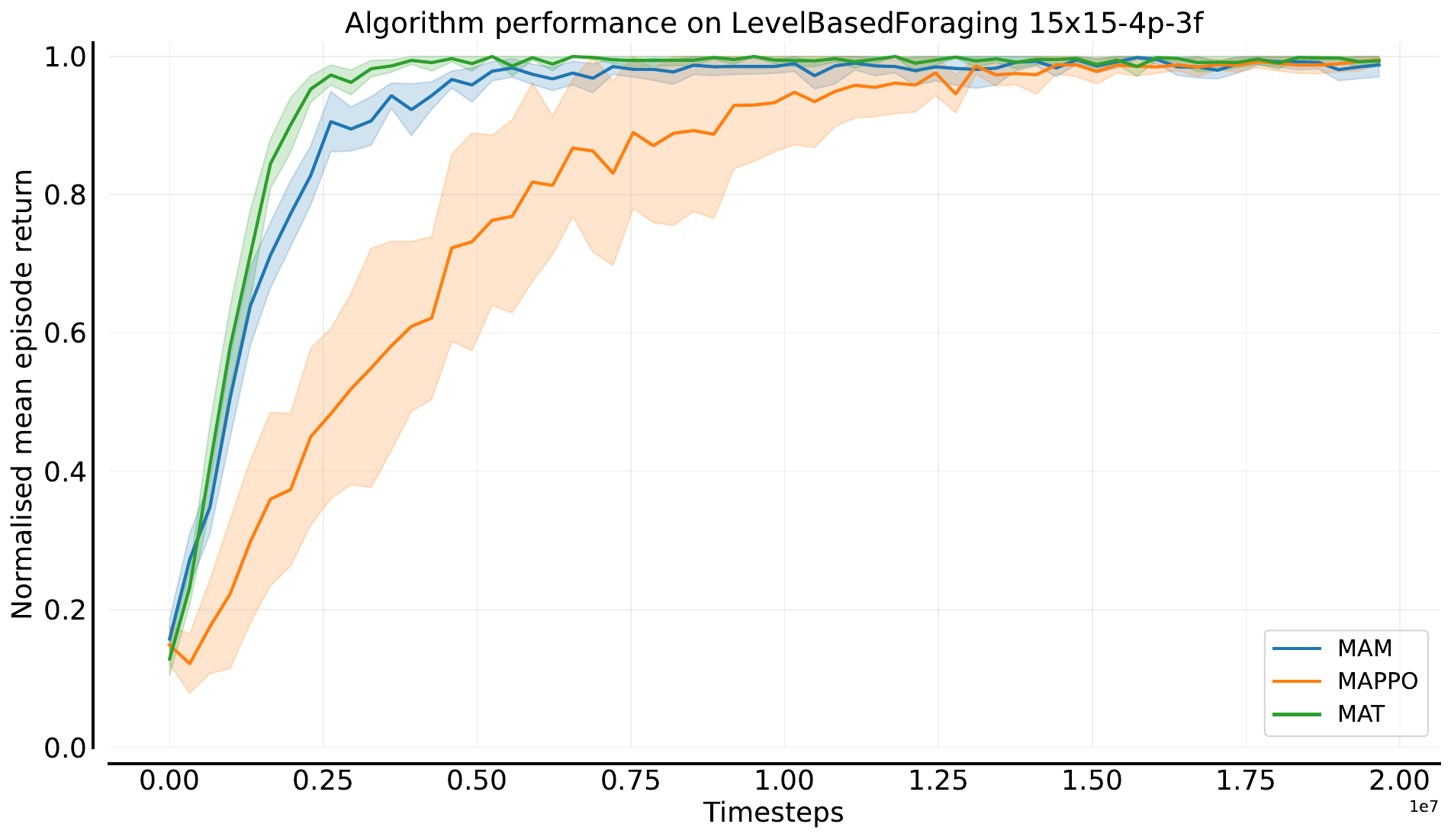}

    \caption{Mean episode return over ten seeds for MAPPO, MAT and MAM with 95\% confidence intervals \citep{agarwal2022deepreinforcementlearningedge} for RWARE, SMAX and LBF tasks.}
    \label{fig:all_task_return_results}
\end{figure}


\begin{table}
    \centering
    \caption{Mean episode return with 95\% confidence intervals for each task.
    Bold values indicate the highest score per task and an asterisk indicates that a score overlaps with the highest score within one confidence interval.}
        \label{tab:indivual_performances}
        \resizebox{\linewidth}{!}{
    \begin{tabular}{ll|ccc} \toprule
    \textit{Environment}            & \textit{Task}      & \textit{MAPPO}       & \textit{MAT}          & \textit{MAM}      \\ \midrule
    \multirow{2}{*}{\textit{RWARE}} & \texttt{tiny-2ag}           & $10.58_{(4.43,16.73)}$          & $17.26_{(15.81,18.71)}*$        & $\mathbf{17.79}_{(17.16,18.43)}$          \\
                                    & \texttt{tiny-4ag}           & $25.45_{(23.42,27.48)}$          & $28.31_{(27.37,29.26)}$        & $\mathbf{29.28}_{(28.06,30.49)}$          \\
                                    & \texttt{small-4ag}          & $11.23_{(10.74,11.72)}$          & $\mathbf{18.60}_{(17.90,19.30)}$        & $16.98_{(15.97,18.00)}$          \\ \midrule

    \multirow{4}{*}{\textit{SMAX}}  & \texttt{2s3z}               & $1.990_{(1.979,2.000)}*$           & $\mathbf{1.993}_{(1.983,2.000)}$         & $1.923_{(1.827,2.000)}*$           \\
                                    & \texttt{3s5z}               & $1.980_{(1.955,2.000)}*$           & $\mathbf{1.997}_{(1.990,2.000)}$         & $1.986_{(1.968,2.000)}*$           \\
                                    & \texttt{3s\_vs\_5z}         & $1.953_{(1.984,1.994)}*$           & $1.952_{(1.926,1.977)}$         & $\mathbf{1.983}_{(1.966,2.000)}$           \\
                                    & \texttt{6h\_vs\_8z}         & $\mathbf{1.994}_{(1.984,2.000)}$           & $1.993_{(1.983,2.000)}*$         & $1.987_{(1.964,2.000)}*$           \\
                                    & \texttt{3s5z\_vs\_3s6z}    & $\mathbf{1.918}_{(1.851,1.985)}$           & $1.916_{(1.872,1.959)}*$         & $1.727_{(1.584,1.869)}$           \\
                                    & \texttt{5m\_vs\_6m}         & $1.072_{(0.690,1.453)}$           & $1.606_{(1.243,1.969)}*$         & $\mathbf{1.726}_{(1.389,2.000)}$           \\
                                    & \texttt{10m\_vs\_11m}       & $1.317_{(1.201,1.433)}$           & $1.867_{(1.731,2.000)*}$         & $\mathbf{1.985}_{(1.966,2.000)}$           \\
                                    & \texttt{27m\_vs\_30m}       & $1.806_{(1.714,1.898)}$           & $1.909_{(1.859,1.959)}$         & $\mathbf{1.961}_{(1.935,1.987)}$           \\
                                    & \texttt{smacv2\_5\_units}   & $1.697_{(1.648,1.747)}$           & $\mathbf{1.788}_{(1.742,1.834)}$         & $1.771_{(1.726,1.816)}*$           \\
                                    & \texttt{smacv2\_10\_units}  & $\mathbf{1.770}_{(1.732,1.807)}$           & $1.635_{(1.554,1.716)}$         & $1.584_{(1.526,1.642)}$           \\
                                    & \texttt{smacv2\_20\_units}  & $\mathbf{1.526}_{(1.444,1.607)}$           & $1.490_{(1.411,1.569)}*$         & $1.477_{(1.421,1.533)}*$           \\ \midrule

    \multirow{4}{*}{\textit{LBF}}   & \texttt{2s-8x8-2p-2f-coop}  & $0.994_{(0.986, 1.000)}*$          & $0.995_{(0.988,1.000)}*$        & $\mathbf{0.998}_{(0.995,1.000)}$              \\
                                    & \texttt{8x8-2p-2f-coop}     & $\mathbf{1.000}_{(1.000,1.000)}$          & $\mathbf{1.000}_{(1.000,1.000)}$        & $\mathbf{1.000}_{(1.000,1.000)}$              \\
                                    & \texttt{2s-10x10-3p-3f}     & $0.850_{(0.812,0.888)}$          & $\mathbf{0.973}_{(0.955,0.991)}$        & $0.972_{(0.949,0.994)}*$              \\
                                    & \texttt{10x10-3p-3f}        & $0.991_{(0.982,1.000)}*$          & $\mathbf{0.997}_{(0.990,1.000)}$        & $\mathbf{0.997}_{(0.991,1.000)}$              \\
                                    & \texttt{15x15-3p-5f}        & $0.620_{(0.501,0.739)}$          & $\mathbf{0.955}_{(0.930,0.980)}$        & $0.954_{(0.931,0.978)}*$              \\
                                    & \texttt{15x15-4p-5f}        & $0.812_{(0.730,0.893)}$          & $\mathbf{0.979}_{(0.967,0.991)}$        & $0.943_{(0.92,0.965)}$              \\
                                    & \texttt{15x15-4p-3f}        & $\mathbf{0.995}_{(0.988,1.000)}$          & $0.994_{(0.985,1.000)}*$        & $0.988_{(0.970,1.000)}*$              \\ \bottomrule
    \end{tabular}
}
\end{table}


\begin{figure}[t]
    \centering
    \begin{subfigure}[b]{0.32\linewidth}
        \centering
        \includegraphics[width=\linewidth]{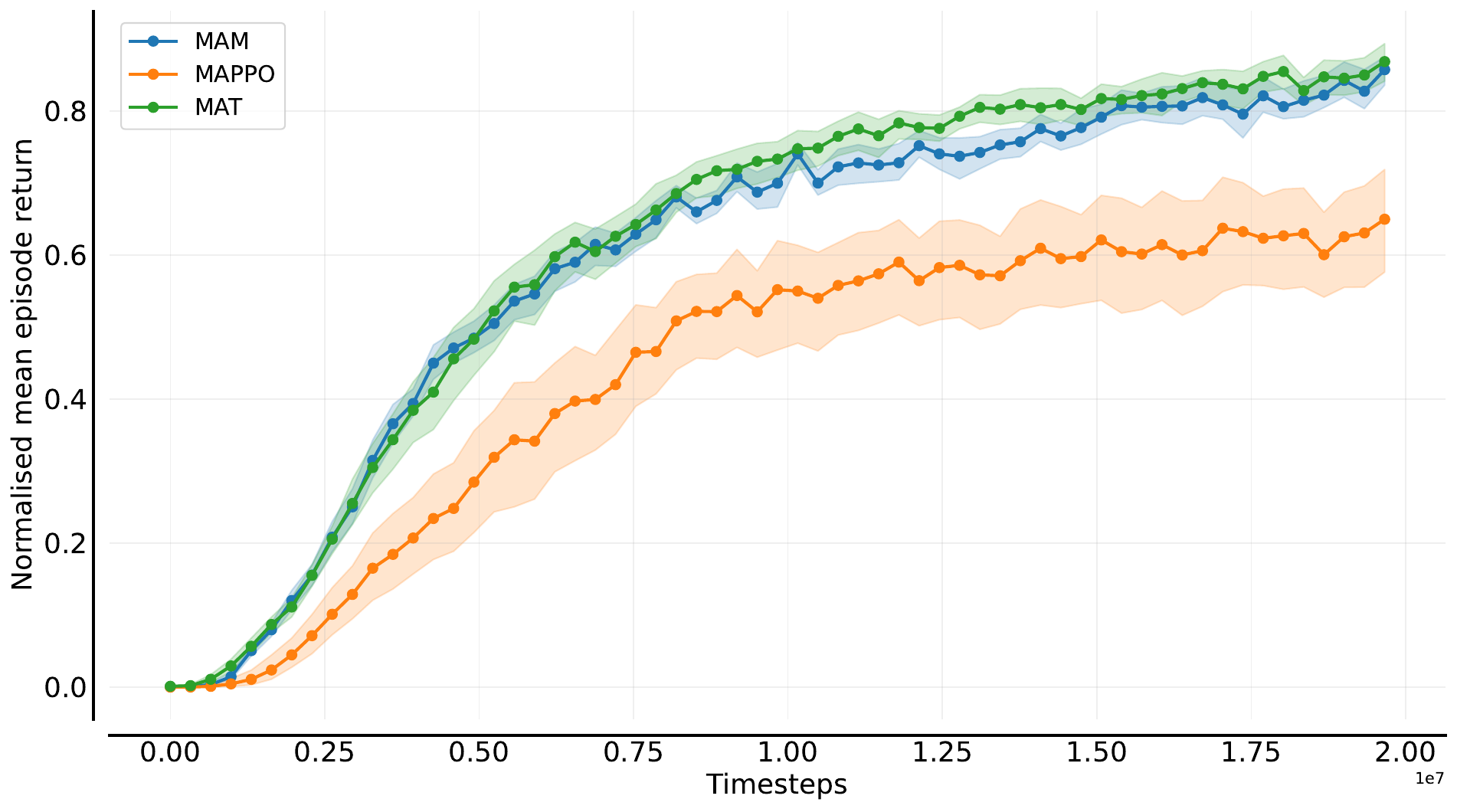}
        \includegraphics[width=\textwidth]{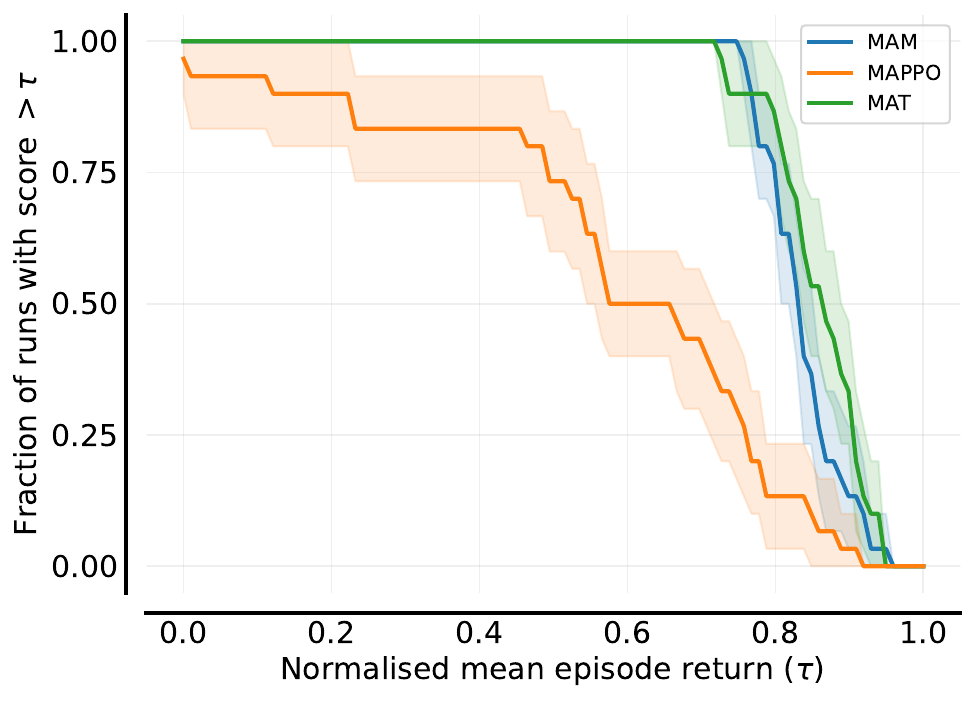}
        \caption{RWARE}
    \end{subfigure}
    \hfill
    \begin{subfigure}[b]{0.32\linewidth}
        \centering
        \includegraphics[width=\linewidth]{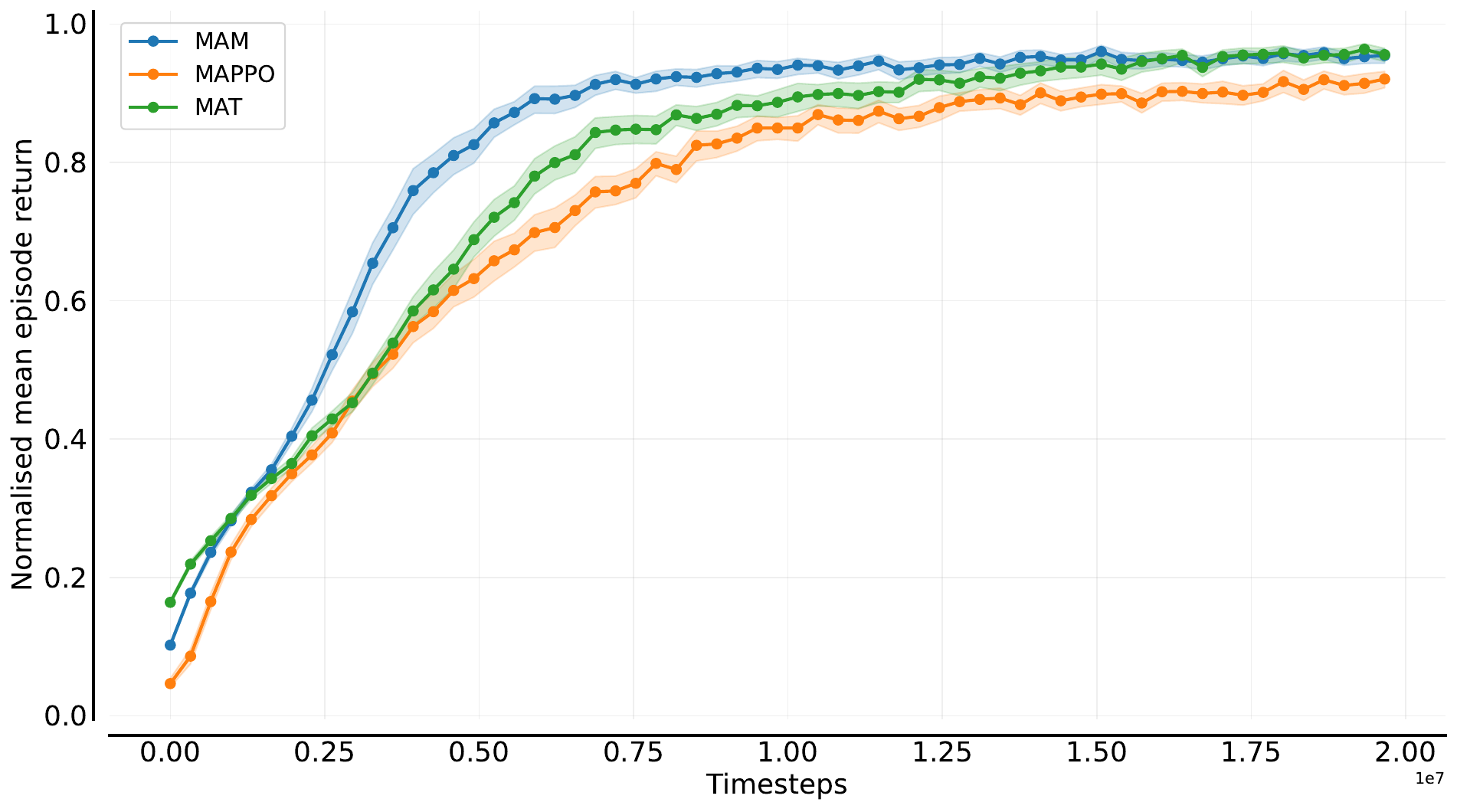}
        \includegraphics[width=\textwidth]{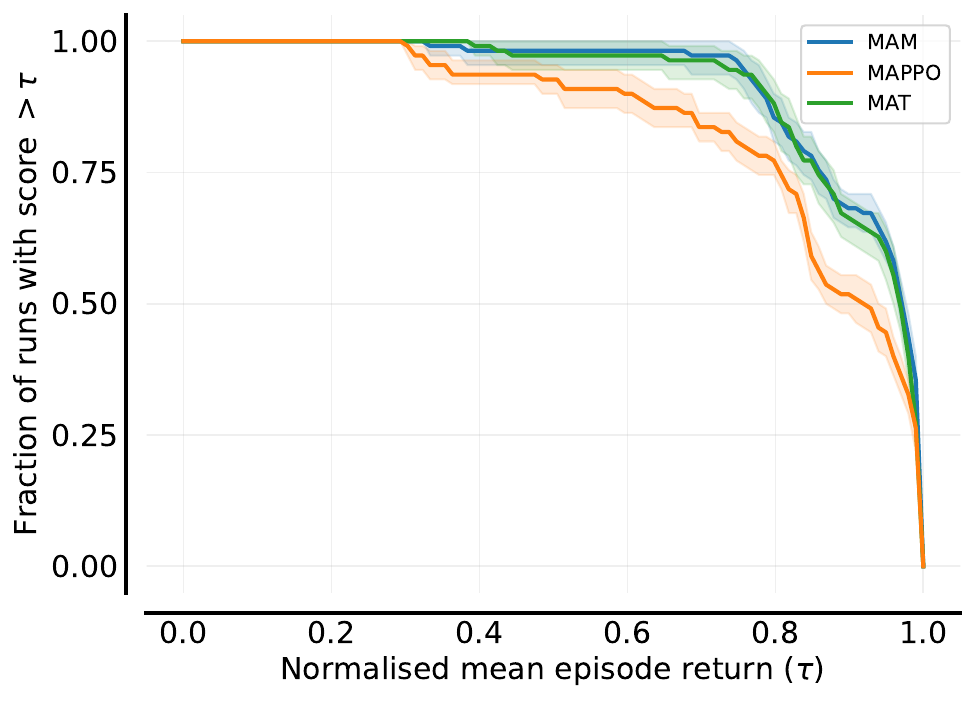}
        \caption{SMAX}
    \end{subfigure}
    \hfill
    \begin{subfigure}[b]{0.32\linewidth}
        \centering
        \includegraphics[width=\linewidth]{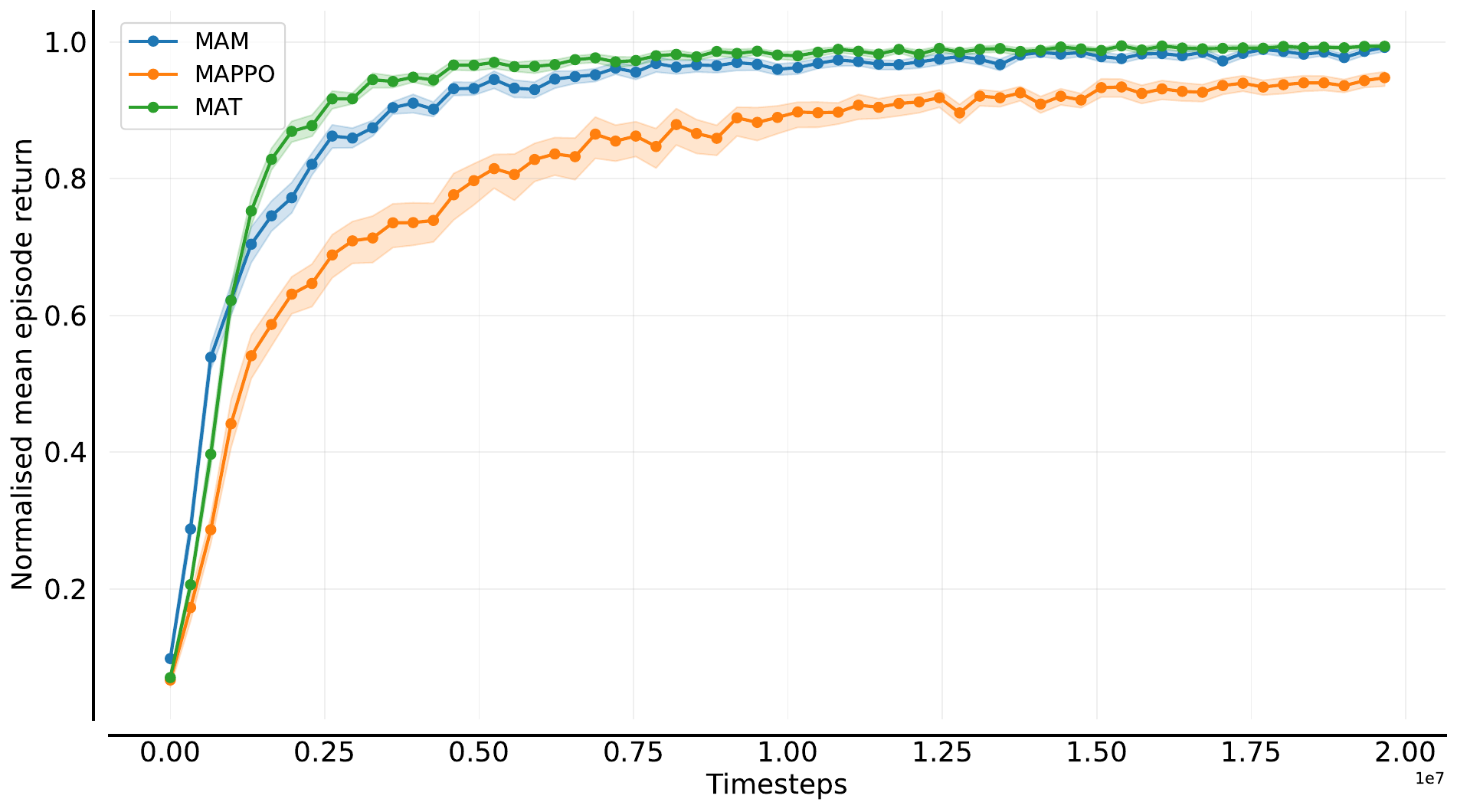}
        \includegraphics[width=\textwidth]{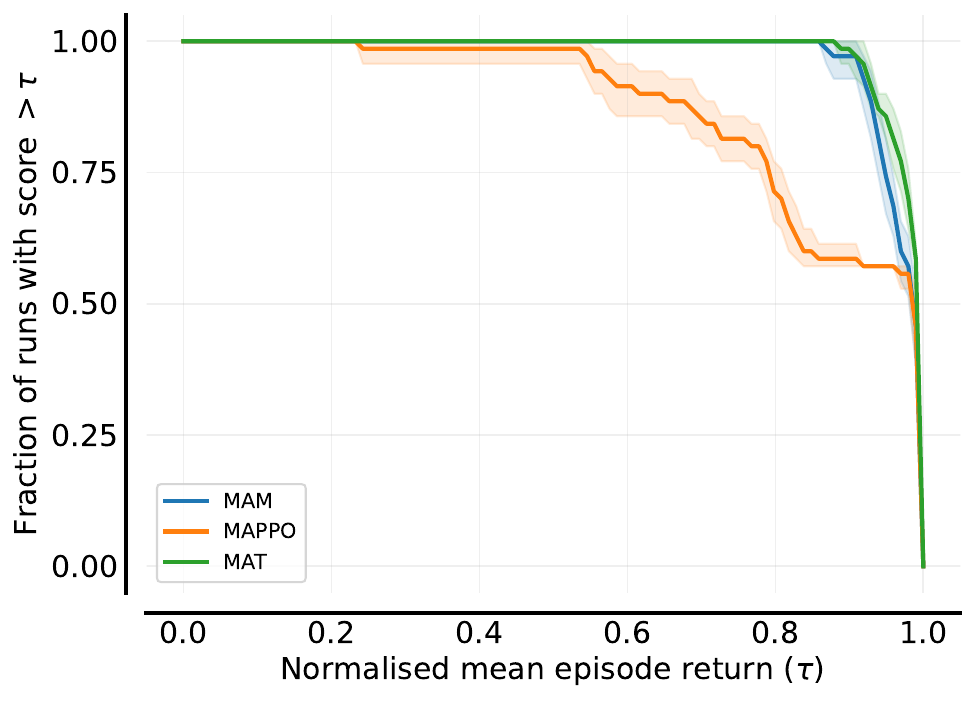}
        \caption{LBF}
    \end{subfigure}

    \caption{Normalised mean episode returns (top) and performance profiles (bottom), aggregated over ten seeds for MAPPO, MAT and MAM, including 95\% bootstrap confidence intervals \citep{agarwal2022deepreinforcementlearningedge}.
    For each of the subplots, the results from all tasks in the corresponding environment were aggregated.
    }
    \label{fig:agg_return_results}
\end{figure}


In this section, we describe our experimental setup, followed by detailed and aggregated results across a wide range of tasks.


\subsection{Experimental Setup}

\paragraph{\textbf{Environments}}
We evaluate MAM on three JAX-based versions of well-established environments: Robotic Warehouse (RWARE) \citep{papoudakis2021benchmarkingmultiagentdeepreinforcement} from Jumanji \citep{bonnet2024jumanjidiversesuitescalable}, Level-Based Foraging (LBF) \citep{christianos2021sharedexperienceactorcriticmultiagent} and the StarCraft Multi-Agent Challenge (SMAX) \citep{rutherford2023jaxmarlmultiagentrlenvironments}.
For each environment, we choose a wide range of tasks with varying difficulty and number of agents.

\paragraph{\textbf{Algorithm Implementation}}
For our implementation, we build on top of the JAX-based MARL library, Mava \citep{dekock2023mava}.
Extra implementation details are included in the supplementary materials.

\paragraph{\textbf{Evaluation Protocol}}
For each task in an environment, we train each algorithm on ten pseudorandom seeds for 20 million timesteps and evaluate at 61 evenly spaced intervals.
At each evaluation, we record the episode return across 32 environment rollouts as recommended by \citet{gorsane2022standardisedperformanceevaluationprotocol}.
In order to aggregate runs across seeds and tasks we use the MARL-eval library \citep{gorsane2022standardisedperformanceevaluationprotocol}.
For task-level aggregation, we report the mean with $95\%$ confidence intervals.
For aggregations over entire environment suites, we report the min-max normalised inter-quartile mean with $95\%$ stratified bootstrap confidence intervals.

\paragraph{\textbf{Hyperparameters}}
All algorithms are tuned on each task with a tuning budget of 40 trials using the Tree-structured Parzen Estimator (TPE) Bayesian optimisation algorithm from the Optuna library \citep{akiba2019optunanextgenerationhyperparameteroptimization}, and limited to similar maximum model sizes.
Additional information regarding the hyperparameters is included in the supplementary materials.


\subsection{Results}

\paragraph{\textbf{Individual Task Results}}
In Figure \ref{fig:all_envs_returns} we show the performance of MAM against MAT and MAPPO aggregated over all environments and task, including 95\% confidence intervals.
MAM matches the aggregated performance of MAT, currently state-of-the-art in the MARL research field, and displays faster learning capabilities.
We find that MAM matches or outperforms MAT in 18 of the 21 tasks, indicating that attention can be replaced without sacrificing performance in MARL tasks.
In Table \ref{tab:indivual_performances} we provide the final mean episode return of each algorithm on each of the tasks, including MAPPO as an additional baseline.
Of all three algorithms, MAM matches or outperforms the highest score with 95\% confidence in 17 of the 21 tasks.
We use bold to highlight the result with the highest score and an asterisk to indicate that a score overlaps with the highest score within one confidence interval.
Individual task plots are shown in Figure \ref{fig:all_task_return_results}.

\paragraph{\textbf{Environment Aggregated Results}}
In order to get a more reliable comparison between the algorithms, and to avoid cherry-picking tasks, we additionally compute aggregated results per environment.
In particular, we first normalise the scores on all tasks in an environment and then aggregate them using MARL-eval \citep{gorsane2022standardisedperformanceevaluationprotocol}.
The aggregated sample efficiency curves and performance profiles are given in Figure \ref{fig:agg_return_results}.
From these plots, we can see that on all three environments MAM achieves very similar final performances to MAT.
Moreover, MAM’s sample efficiency closely matches MAT’s on RWARE and LBF.
Interestingly, on SMAX MAM’s sample efficiency is significantly better than that of MAT.
This is noteworthy because SMAX tasks feature an average of 9.7 agents, compared to 3 and 3.3 on LBF and RWARE, respectively.
This suggests that, on tasks with more agents, MAM may have superior sample-efficiency.
However, more experimental evidence on tasks with even larger numbers of agents is required to verify this claim.

\paragraph{\textbf{Inference Speed Results}}
Due to its selectivity, Mamba executes in only recurrent mode.
This proves advantageous at inference-time, shown in Figure \ref{fig:inference_sps_results}.
Empirically, we observe that MAM scales better than MAT in number of agents.

This result aligns with existing scaling results \citep{vaswani2023attentionneed,gu2024mambalineartimesequencemodeling}.
Faster inference time could allow real-world, many-agent systems to be better modelled by MAM than MAT--the latter of which will become prohibitively slow.



\section{Model Ablations}

The inclusion of Mamba in the MAT architecture introduces two Mamba-specific parameters: the size of the hidden state dimension, $N$, and the projection dimension of $\bm{\Delta}$.
The parameter $\bm{\Delta}$ controls the amount of information included from the current input, and generalises the RNN gating mechanism.
We conduct an ablation study over the projection dimension of $\bm{\Delta}$ and the hidden state dimension $N$, shown in Tables \ref{tab:delta_rank_ablation} and \ref{tab:latent_dim_ablation}, respectively.
We conduct ablations on RWARE because, unlike SMAX and LBF, this environment has no reward ceiling, allowing for clearer differentiation between different results.
We use the RWARE task $\texttt{small-4ag}$ to assess hyperparameter impacts on the greatest number of agents possible within our set of RWARE benchmarking tasks.
In contrast to the findings of \citet{gu2024mambalineartimesequencemodeling}, our results show that increasing the dimension size for $\bm{\Delta}$ and $N$ is not always advantageous.
For both $\bm{\Delta}$ and $N$, we achieve the best episode return and smallest variation between runs at the smallest hyperparameter value.
Episode return trends downwards while standard deviation trends upwards as the size of $N$ or the projection dimension of $\bm{\Delta}$ becomes large.
We hypothesise that existing MARL benchmarking environments contain too few agents to observe a marked performance improvement when increasing information flow through $\bm{\Delta}$ or the hidden state via $N$.
Small-to-modest values for these hyperparameters allow sufficient communication between agents to solve these tasks and stabilise training, and larger values risk adding unnecessary noise.
The balance between information compression and performance may become more apparent in significantly many-agent environments.
Using smaller hyperparameter values for $\bm{\Delta}$ and $N$, a lightweight MAM model could prove both effective and efficient across current MARL benchmarks.


 \begin{table}[t]
    \centering
        \caption{Varying the projection dimension for $\bm{\Delta}$--a selection mechanism of Mamba--on RWARE's \texttt{small-4ag} task.
        Aggregated mean episode return and corresponding standard deviations are computed over ten independent runs.
        We hold $N$ at 32.}
        \label{tab:delta_rank_ablation}
    \begin{tabular}{c c c c}
        \toprule
        $\bm{\Delta}$ & \textit{Parameter Count} & \textit{Episode Return} & \textit{Standard Devation} \\ \midrule
        1 & 484882 & 12.65 & 1.24 \\
        2 & 486418 & 11.76 & 2.56 \\
        4 & 489490 & 11.82 & 2.75 \\
        8 & 495634 & 11.89 & 2.76 \\
        16 & 507922 & 12.10 & 2.01 \\
        32 & 532498 & 12.61 & 1.94 \\
        64 & 581650 & 12.16 & 2.25 \\
        128 & 679954 & 9.72 & 5.22 \\ \bottomrule
    \end{tabular}
\end{table}


 \begin{table}[t]
    \centering
        \caption{Varying hidden dimension size on RWARE's \texttt{small-4ag} task.
    Aggregated mean episode return and corresponding standard deviations are computed over independent ten runs.
    We hold the projection dimension of $\bm{\Delta}$ at 128.}
        \label{tab:latent_dim_ablation}
    \begin{tabular}{c c c c}
        \toprule
        $N$ & \textit{Parameter Count} & \textit{Episode Return} & \textit{Standard Deviation} \\ \midrule
        1 & 612498 & 12.26 & 1.61 \\
        2 & 614674 & 12.05 & 2.39 \\
        4 & 619026 & 10.48 & 3.53 \\
        8 & 627730 & 10.27 & 4.25 \\
        16 & 645138 & 10.33 & 3.03 \\
        32 & 679954 & 11.35 & 4.15 \\ \bottomrule
    \end{tabular}
\end{table}



\section{Discussion}

\paragraph{\textbf{Limitations}}

In this paper, we have not explored the effectiveness of MAM in tasks involving very large numbers of agents.
This constraint arises because such tasks are currently limited in the MARL literature.
Naively scaling agent numbers in LBF and SMAX results in sharp rises in time and memory cost since the size of the observation space increases with the number of agents.
RWARE offers control over observation space size through modification of agents' field of view, but this also substantially increases its difficulty, preventing informative benchmarking on a many-agent modification.

Additionally, MAM and MAT are implemented using JAX to leverage the accelerated MARL library Mava \citep{dekock2023mava}.
Although JAX's Accelerated Linear Algebra (XLA) compiler automatically optimises code at runtime for faster execution on various hardware--including CPUs, GPUs, and TPUs--we cannot replicate the hardware-aware parallelism achieved by the original Mamba implementation in PyTorch, which uses custom CUDA kernels.
JAX's tool for writing custom kernels, Pallas, is relatively new and still under active development.
While promising, Pallas has not yet reached the same level of stability, performance optimisation, and widespread adoption as CUDA.
Moreover, JAX abstracts away some of the complexity required for highly optimised, architecture-specific code, which may result in CUDA outperforming Pallas for cutting-edge performance optimisations.
Consequently, our current MAM implementation cannot match the speed and memory benefits of the original implementation during training.

\paragraph{\textbf{Future Work}}

To address these limitations, we plan to create or modify environments to facilitate the evaluation of MAM in many-agent tasks.
Developing new environments or adapting existing ones will enable more informative benchmarking and exploration of MAM's capabilities in settings with a large number of agents.

Furthermore, we intend to implement MAM in PyTorch to leverage the performance benefits of custom CUDA kernels, as demonstrated in the original Mamba implementation.
We believe that such an implementation will offer equivalent performance and improved computational costs, especially when used in well-implemented environments designed for many-agent tasks.



\section{Related Work}

Two existing works are related to MAM: Decision Mamba \citep{ota2024decisionmambareinforcementlearning} for offline single-agent RL, and CrossMamba for target sound extraction \citep{wu2024crossattentioninspiredselectivestate}.

Decision Mamba modifies the Decision Transformer \citep{chen2021decisiontransformerreinforcementlearning} by replacing causal self-attention with a vanilla Mamba block.
Designed for offline single-agent RL, Decision Mamba uses the agent’s trajectory over time as its input sequence.
In contrast, MAM is tailored for online learning in a multi-agent setting.
It processes the observations of multiple agents at each timestep as its input sequence and sequentially decodes actions for each agent.
Additionally, MAM not only replaces causal self-attention but also provides alternatives to non-causal self-attention and cross-attention.

The concurrent work by \citet{wu2024crossattentioninspiredselectivestate} introduces the CrossMamba block for extracting target sounds from audio features, serving as a cross-attention replacement in the AV-SepFormer and Waveformer architectures.
While similar to the cross-attention Mamba block we propose, CrossMamba was developed independently for an entirely different domain.



\section{Conclusion}

In this work, we introduced the Multi-Agent Mamba, a new sequence-based architecture for MARL that matches the performance of existing state-of-the-art methods with improved efficiency for many-agent scenarios.
We substitute causal self-attention for the vanilla Mamba block, non-causal self-attention for a bi-directional Mamba block variant, and causal cross-attention for a novel `cross-attentional' Mamba block adapted to operate on two input sequences.
This new model offers fast inference and linear scaling as the number of agents increases, without compromise on performance.
We demonstrate its performance against MAT, currently state-of-the-art, and MAPPO on a range of well-known MARL benchmarking environments.
This suggests that Mamba can port well to MARL tasks, solving scaling issues in tasks with many agents.
It is our hope that this investigation contributes to the further development of efficient and effective multi-agent decision-making algorithms.



\newpage
\bibliographystyle{plainnat}
\bibliography{references}


\newpage

\appendix


\section{Implementation Details}

We closely follow the original Mamba authors' \citep{gu2024mambalineartimesequencemodeling} implementation, which includes some modifications to the SSM theoretical results laid out in this paper.

The first of these is the addition of a residual connection to the SSM equations in \ref{eqn:cont_SSM}, which becomes
\begin{equation}
    \begin{aligned}
        h'(t) &= \mathbf{A}h(t) + \mathbf{B}x(t) \label{eqn:cont_SSM_with_D} \\
        y(t) &= \mathbf{C}h(t) + \mathbf{D}x(t),
    \end{aligned}
\end{equation}
where all variables are defined the same as in \ref{eqn:cont_SSM} and $\mathbf{D} \in \mathbb{R}^{D \times D}$ can be viewed as a residual connection that allows the current input to further influence the SSM output.

Second is the use of a first-order approximation when discretising $\mathbf{B}$, particularly $\mathbf{\bar{B}}=\bm{\Delta}\mathbf{B}$.
This corresponds to using the Euler discretisation method for $\mathbf{B}$.
Given \ref{eqn:disc_SSM_params} and \ref{eqn:selective_SSM_params}, but using this approximation for $\mathbf{B}$, we can instead rewrite \ref{eqn:mamba_attn_element} as
\begin{equation}
    \begin{aligned}
        \Lambda_{ij}^* &= S_\mathbf{C}(x_i) \left( \prod^i_{k=j+1} \exp\left(\text{softplus}\left(S_{\bm{\Delta}}(x_k)\right) \mathbf{A}_k \right) \right) \\
        &\quad \Bigl( \text{softplus} \left( S_{\bm{\Delta}}(x_j) \right) S_\mathbf{B}(x_j) \Bigr) \\
        &= \mathbf{\tilde{Q}}_i \mathbf{H}^*_{i,j} \mathbf{K}^*_j, \label{eqn:mamba_approx_attn_element_QHK}
    \end{aligned}
\end{equation}
where $\mathbf{\tilde{Q}}$ remains the equivalent query matrix, $\mathbf{K}^*$ is the key matrix, and $\mathbf{H}^*$ is an additional matrix unique to Mamba that again summarises the historical context between sequence tokens $x_j$ to $x_i$.
Importantly for our purposes, this does not change cross-attention formulation for CrossMamba.


\section{Hyperparameter Settings}

We utilise Mava's design architecture to distribute the entire reinforcement learning training loop across multiple devices with the JAX \texttt{pmap} transformation, in addition to vectorising training with the JAX \texttt{vmap} transformation.

\subsection{Benchmarking Results}

All algorithms on all tasks are run using 64 vectorised environments.
We use the same default hyperparameters and hyperparameter search spaces for a given algorithm on all tasks.
We constrain the maximum model size allowable during hyperparameter tuning similarly for both MAM and MAT.
All algorithms are tuned using the TPE Bayesian optimisation algorithm from the Optuna library \citep{akiba2019optunanextgenerationhyperparameteroptimization} with a tuning budget of 40 trials.
The default hyperparameters are given in Tables \ref{tab:MAM_default_hyperparams}, \ref{tab:MAT_default_hyperparams} and \ref{tab:MAPPO_default_hyperparams}, and the discrete search spaces for each algorithm are given in Tables \ref{tab:MAM_search_space}, \ref{tab:MAT_search_space} and \ref{tab:MAPPO_search_space}.


\begin{table}
    \centering
    \caption{Default hyperparameters for MAM.}
        \label{tab:MAM_default_hyperparams}
    \begin{tabular}{c c} \toprule
        \textit{Hyperparameter} & \textit{Value} \\ \midrule
        Activation function & GeLU \\
        Normalise advantage & False \\
        Value function coefficient & 0.5 \\
        Discount $\gamma$ & 0.99 \\
        GAE $\lambda$ & 0.9 \\
        Entropy coefficient & 0.01 \\
        Rollout length & 128 \\
        Add one-hot agent ID & True \\
        Number of Mamba blocks & 1 \\
        1D convolution dimension & 4 \\ \bottomrule
    \end{tabular}
\end{table}


\begin{table}
    \centering
    \caption{Default hyperparameters for MAT.}
        \label{tab:MAT_default_hyperparams}
    \begin{tabular}{c c} \toprule
        \textit{Hyperparameter} & \textit{Value} \\ \midrule
        Activation function & GeLU \\
        Normalise advantage & False \\
        Value function coefficient & 0.5 \\
        Discount $\gamma$ & 0.99 \\
        GAE $\lambda$ & 0.9 \\
        Entropy coefficient & 0.01 \\
        Rollout length & 128 \\
        Add one-hot agent ID & True \\ \bottomrule
    \end{tabular}
\end{table}


\begin{table}
    \centering
    \caption{Default hyperparameters for MAPPO.}
        \label{tab:MAPPO_default_hyperparams}
    \begin{tabular}{c c} \toprule
        \textit{Hyperparameter} & \textit{Value} \\ \midrule
        Critic network sizes & $[128,128]$ \\
        Policy network sizes & $[128,128]$ \\
        Activation function & RelU \\
        Normalise advantage & False \\
        Value function coefficient & 0.5 \\
        Discount $\gamma$ & 0.99 \\
        GAE $\lambda$ & 0.9 \\
        Entropy coefficient & 0.01 \\
        Rollout length & 128 \\
        Add one-hot agent ID & True \\ \bottomrule
    \end{tabular}
\end{table}


\begin{table}
    \centering
    \caption{Hyperparameter search space for MAM.}
        \label{tab:MAM_search_space}
    \begin{tabular}{c c} \toprule
        \textit{Hyperparameter} & \textit{Value} \\ \midrule
        PPO epochs & $\{2,5,10\}$ \\
        Number of minibatches & $\{1,2,4,8\}$ \\
        Clipping $\epsilon$  & $\{0.05,0.1,0.2\}$ \\
        Maximum gradient norm & $\{0.5,5,10\}$ \\
        Learning rate & $\{0.0001,0.00025,0.0005,0.001\}$ \\
        Model embedding dimension & $\{32,64,128\}$ \\
        Hidden state dimension & $\{1,2,4,8,16,32\}$ \\
        $\bm{\Delta}$ projection dimension & $\{1,2,4,8,16,32,64,128\}$ \\ \bottomrule
    \end{tabular}
\end{table}


\begin{table}
    \centering
    \caption{Hyperparameter search space for MAT.}
        \label{tab:MAT_search_space}
    \begin{tabular}{c c} \toprule
        \textit{Hyperparameter} & \textit{Value} \\ \midrule
        PPO epochs & $\{2,5,10\}$ \\
        Number of minibatches & $\{1,2,4,8\}$ \\
        Clipping $\epsilon$  & $\{0.05,0.1,0.2\}$ \\
        Maximum gradient norm & $\{0.5,5,10\}$ \\
        Learning rate & $\{0.0001,0.00025,0.0005,0.001\}$ \\
        Model embedding dimension & $\{32,64,128\}$ \\
        Number of Transformer heads & $\{1,2,4\}$ \\
        Number of Transformer blocks & $\{1,2,3\}$ \\ \bottomrule
    \end{tabular}
\end{table}


\begin{table}
    \centering
    \caption{Hyperparameter search space for MAPPO.}
        \label{tab:MAPPO_search_space}
    \begin{tabular}{c c} \toprule
        \textit{Hyperparameter} & \textit{Value} \\ \midrule
        PPO epochs & $\{2,4,8\}$ \\
        Number of minibatches & $\{2,4,8\}$ \\
        Entropy coefficient & $\{0,0.01,0.00001\}$ \\
        Clipping $\epsilon$  & $\{0.05,0.1,0.2\}$ \\
        Maximum gradient norm & $\{0.5,5,10\}$ \\
        Critic learning rate & $\{0.0001,0.00025,0.0005\}$ \\
        Policy learning rate & $\{0.0001,0.00025,0.0005\}$ \\ \bottomrule
    \end{tabular}
\end{table}


\subsection{Inference Speed Results}

MAT and MAM are run for 30 evaluations of 32 episodes and averaged over three seeds, and we use two parallel environments.
We plot the inverted mean steps per second during evaluation.
The hyperparameter settings are shown in Tables \ref{tab:MAM_inference_hyperparams} and \ref{tab:MAT_inference_hyperparams}.
We use NVIDIA A100 80GB GPUs to conduct these experiments.


\begin{table}
    \centering
    \caption{Hyperparameter values for MAM inference speed tests.}
        \label{tab:MAM_inference_hyperparams}
    \begin{tabular}{c c} \toprule
        \textit{Hyperparameter} & \textit{Value} \\ \midrule
        PPO epochs & 10 \\
        Number of minibatches & 2 \\
        Clipping $\epsilon$  & 0.1 \\
        Maximum gradient norm & 0.5 \\
        Learning rate & 0.0005 \\
        Model embedding dimension & 128 \\
        Hidden state dimension & 32\\
        $\bm{\Delta}$ projection dimension & 128 \\ \bottomrule
    \end{tabular}
\end{table}


\begin{table}
    \centering
    \caption{Hyperparameter search space for MAT inference speed tests.}
        \label{tab:MAT_inference_hyperparams}
    \begin{tabular}{c c} \toprule
        \textit{Hyperparameter} & \textit{Value} \\ \midrule
        PPO epochs & 10 \\
        Number of minibatches & 2 \\
        Clipping $\epsilon$  & 0.1 \\
        Maximum gradient norm & 0.5 \\
        Learning rate & 0.0005 \\
        Model embedding dimension & 128 \\
        Number of Transformer heads & 1 \\
        Number of Transformer blocks & 3 \\ \bottomrule
    \end{tabular}
\end{table}


\subsection{Ablation Results}

We run ablation experiments using the same experimental setup as for the benchmarking results.
The values of the hyperparameters held constant during ablations are listed in Table \ref{tab:MAM_ablation_hyperparams}.


\begin{table}
    \centering
    \caption{Values of hyperparameters held constant for MAM ablations.}
        \label{tab:MAM_ablation_hyperparams}
    \begin{tabular}{c c} \toprule
        \textit{Hyperparameter} & \textit{Value} \\ \midrule
        PPO epochs & 2 \\
        Number of minibatches & 1 \\
        Clipping $\epsilon$  & 0.1 \\
        Maximum gradient norm & 5 \\
        Learning rate & 0.001 \\
        Model embedding dimension & 128 \\ \bottomrule
    \end{tabular}
\end{table}


\section{Computational Resources}

Model training and inference was run on various GPUs, namely NVIDIA Quadro RTX 4000 (8GB), Tesla V100 (32GB), and NVIDIA A100 (40GB and 80GB).



\end{document}